%% file: main_arxiv.tex
\documentclass{article} 
\usepackage{iclr2027_conference,times}
\input{math_commands.tex}

\usepackage{hyperref}
\usepackage{url}
\usepackage{graphicx}
\usepackage{booktabs}
\usepackage{amsmath}
\usepackage{multirow}

\newcommand{\msun}{mSUN}
\newcommand{\ehull}{E_{\mathrm{hull}}}
\newcommand{\priorrate}{13.4\%}      
\newcommand{\headlinerate}{45.5\%}   
\newcommand{\gainx}{3.4$\times$}

\title{Reinforcement Learning on the Discrete Composition Channel of a
Crystal Generator: Validated Gains and Reward Hacking}

\author{%
Pawan Prakash$^{1}$, Philipp H\"ollmer$^{3}$, Addis Fuhr$^{2}$, Peter Hirschfeld$^{1}$,\\
\textbf{P.~Ganesh$^{2}$, Stefano Martiniani$^{3}$, Richard Hennig$^{1}$\thanks{Corresponding author: \texttt{rhennig@ufl.edu}}}\\
$^{1}$University of Florida\quad
$^{2}$Oak Ridge National Laboratory\quad
$^{3}$New York University
}

\iclrfinalcopy
\makeatletter
\let\arxiv@iclrmaketitle\@maketitle
\def\@maketitle{\begingroup\let\lhead\@gobble\arxiv@iclrmaketitle\endgroup}
\makeatother
\begin{document}
\maketitle

\begin{abstract}
Inverse materials design is a long-standing goal of computational materials discovery. Generative models for crystalline materials are typically trained to match the distribution of a structure database, while nothing in their training objective points them at specific design goals such as targeted properties.
We use group-relative policy optimization (GRPO) to align a generative model based on stochastic interpolants and discrete flow matching with general black-box reward functions through reinforcement learning.
Atom types are generated by a discrete flow and the policy gradient of our generalization of GRPO directly acts on the likelihoods of the atom-type transitions, which differentiates our work from previous reinforcement-learning approaches for diffusion and flow-based generative models of crystalline materials.
We introduce a reward function that raises the yield of metastable, unique and novel structures (mSUN) from 13.4\% for the pretrained model to 45.5\% for the reinforced model, as evaluated by a community benchmark.
Our reward also improves the performance of a reinforcement learning framework for crystalline materials based on latent denoising diffusion models. At the same time, we find that directly reinforcing atom-type transition likelihoods enables reward exploitation that has to be prevented with explicit guards. The same analysis also exposes a gap in the community metric. Single-element structures in distinct packings are counted as metastable, unique and novel materials and inflate mSUN without yielding any new compounds.
A stability claim is only as good as its reference hull. We report every result split by the number of reference phases behind it and argue that benchmarks should do the same.
\end{abstract}

\section{Introduction}\label{sec:intro}

Discovering crystalline materials with target properties is a central goal of materials science, and machine learning now accelerates that search at every stage \citep{gnome,mattergen,gibson2026workflow,albavera2025spincrossover}. Generative models trained on databases of experimental and computationally relaxed structures \citep{cdvae,diffcsp,flowmm,mattergen,omatg} learn the distribution of known materials, so they learn stability only implicitly. The field usually evaluates them by the fraction of generated samples that are stable, unique and novel \citep{mattergen,lematgenbench,szymanski2025baselines}, but the training objective itself rewards neither stability nor any other design goal. Conditional generation and classifier-free guidance \citep{hosalimans2022cfg} can steer a model toward a property \citep{unimat,mattergen,crystalgrw,Prakash2026}, but they require property labels for the training structures. Reinforcement learning, in contrast, requires only a reward that can be evaluated on a generated structure, such as an energy from a machine-learned interatomic potential \citep{macemp0,uma,prakash2026uf3}. It has become a common way to reinforce diffusion and flow models toward such rewards \citep{ddpo,dpok,flowgrpo}, with group-relative policy optimization (GRPO) \citep{deepseekmath} as a particularly frequent choice in recent work. Crystal generators have been reinforced this way as well, but in diffusion and flow-based generators the atom type transitions have so far not been treated as separate, discrete actions of the policy. They are either frozen during reinforcement, relaxed to continuous variables, or decoded after the fact from a latent space \citep{omatgirl,crystalgrpo,matinvent,chemeleon2rl}. This leaves a particularly important part of crystal generation, the elemental constituents of a structure that decide its chemistry, only indirectly controlled by the reward. A reward for stable new compounds is largely a reward for choosing favorable combinations of elements, but whether directly reinforcing these compositional decisions improves generation performance remains an open question.

\begin{figure}[t]
\centering
\includegraphics[width=\linewidth]{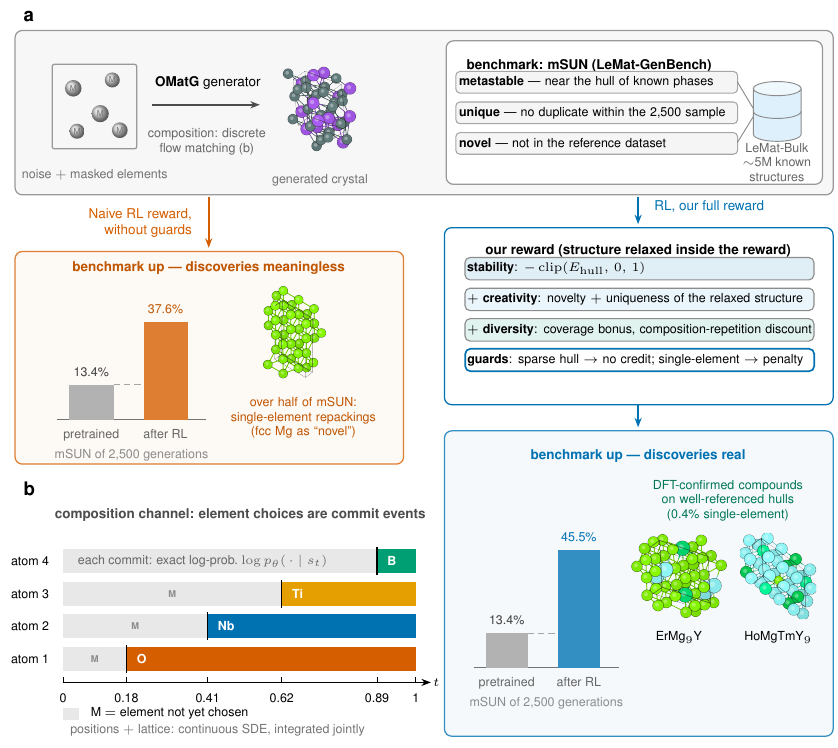}
\caption{\textbf{Reinforcement learning on a crystal generator can satisfy the benchmark without real discoveries, but guards in the reward close the exploit.} \textbf{(a)} The OMatG generator turns noise with masked element identities into crystals and is scored by mSUN under LeMat-GenBench (Section~\ref{sec:setup}). Without the single-element guard, RL inflates the pretrained model's 13.4\% to 37.6\%, but over half of that mSUN set is single-element repackings of known elements, such as face-centered-cubic Mg counted as a novel material. With the full reward (right, Section~\ref{sec:reward}), which routes sparse-hull structures to abstention and single-element structures to the penalty reward, mSUN reaches 45.5\% with 0.4\% single-element structures and DFT-confirmed compounds on well-referenced hulls. \textbf{(b)} The composition channel. Under masked discrete flow matching, each atom's element identity is a discrete commit event with an exact log-probability, while positions and lattice evolve as a continuous SDE (Section~\ref{sec:recipe}).}
\label{fig:concept}
\end{figure}

We reinforce OMatG \citep{omatg} with GRPO \citep{deepseekmath}. OMatG treats the continuous positions and lattice through stochastic interpolants \citep{albergo2023} and the discrete atom types by masked discrete flow matching \citep{campbell2024}. Each atom type is chosen in a single categorical draw at one time step during inference (Fig.~\ref{fig:concept}b), and the only model-dependent factor in that draw is the predicted element distribution. The likelihood ratio has a closed form and the policy gradient reaches the composition channel exactly (Section~3.2). Our reward relaxes every generated structure with the machine-learned potential UMA \citep{uma} and scores its stability by the energy above the convex hull of known competing phases, its novelty and uniqueness, and its contribution to the diversity of the batch, with guards for hulls built from only a few known phases and for single-element structures (Section~3.3). Under the community benchmark LeMat-GenBench \citep{lematgenbench}, the fraction of generated structures that is metastable, unique and novel (\msun{}) rises from 13.4\% for the pretrained model to 45.5\% after reinforcement (Fig.~\ref{fig:concept}a). In contrast, selecting the best of many samples from the pretrained model reaches 7.7\%, and reinforcement learning with the composition generator frozen only reaches 9.8\%. This shows that large improvements require reinforcement learning on the composition channel. Density functional theory agrees with the machine-learned potential on 63 of 67 spot-checked structures. We also consider Chemeleon2 \citep{chemeleon2rl} which reinforces a latent diffusion model whose atom types are decoded from the latent by a frozen decoder, so it changes the composition only indirectly. The mSUN metric of Chemeleon2's released model reaches 36.6\%, and our reward raises it to 43.7\% when it replaces that model's own reward inside its pipeline. Our reward thus appears generally favorable, but the best overall performance is achieved in OMatGRPO with separate, explicitly reinforced atom type transitions.

Each term and guard of our reward was added after the reinforced policy had exploited a previous version, scoring well without producing better materials \citep{skalse}. For instance, an unbounded stability term drove the policy into compositions that the machine-learned potential artificially places far below a convex hull with almost no competing phases. In another exploit, the reinforced policy converged to repack single elements into new unit cells (Fig.~\ref{fig:concept}a). Such a new packing of a known element is not a relevant new material. It still relaxes to an energy near the hull of that element, which holds only a few known competing phases, and it counts as unique and novel because no reference structure matches that packing. Such structures therefore pass every filter of the mSUN metric in LeMat-GenBench, in MatterGen's evaluation code and in Chemeleon2's definition, and they made up 53.5\% of the unguarded run's mSUN set (Section~4.3). This also highlights the flaw in the community benchmark itself. Any stability label is only as good as the hull it is measured against, so we report the number of reference phases behind every generated structure and tier our results by it (Section~4.5).

In summary, this work makes three contributions. To our knowledge, it is the first to apply GRPO to the discrete flow-matching composition channel of a crystal generator, using exact per-event likelihoods so that the elements of a structure are actions of the policy in their native discrete form (Section~3.2). It introduces a reward that more than triples the benchmark yield of the generator, survives selection and frozen-composition controls as well as DFT spot-checks, and transfers to another published generator (Sections~3.3 and 4.2--4.4). Finally, it also records how the policy exploited each earlier version of the reward, and shows that single-element structures and hulls with few reference phases can inflate a community benchmark (Sections~4.3 and 4.5).


\section{Related work}
\label{sec:related}

Most generative models for inorganic crystals learn the distribution of known materials \citep{materialsproject,alexandria} by score-based diffusion \citep{cdvae,diffcsp,mattergen} or flow matching \citep{lipman2023,flowmm}. OMatG \citep{omatg} builds on stochastic interpolants \citep{albergo2023} for the continuous atom positions and lattice vectors, and masked discrete flow matching for the atom types \citep{campbell2024,gat2024dfm}. During inference, an element is chosen in a discrete jump rather than denoised as a continuous variable. Policy-gradient reinforcement learning, which is widely adopted for image diffusion and flow models \citep{ddpo,dpok,flowgrpo,dancegrpo,deepseekmath}, was enabled in OMatG-IRL \citep{omatgirl} for the stochastic interpolants of OMatG, which treat the atom positions and lattice vectors. OMatG-IRL, however, kept the species-generation frozen during reinforcement learning. A discrete jump needs its transition probability, and such policy gradients exist for masked diffusion language models \citep{sepo,d1grpo,maskgrpo} and for discrete flow matching over graphs and sequences \citep{graphgrpo,domino,dflowgrpo}.

Reinforcement learning has reached crystals at several levels of representation. The earliest agents acted on the composition, choosing elements for a formula or for the sites of a fixed template \citep{pan2022rl,crystalgym,crystalgfn}. Language models that write a crystal as text were then reinforced with policy gradients or preference optimization \citep{crystalformerrl,hong2025llmrl,crystalreasoner,plaidpp,crystalicml2026}, which reaches every token including the elements, although some of these works fix the formula in the prompt. The most recent step is reinforcement learning of the diffusion and flow generators that produce complete structures in three dimensions. In these models the composition has not entered the policy gradient as a discrete variable. It is frozen as a condition \citep{omatgirl,crystalgrpo,packflow}, relaxed to continuous variables \citep{matinvent}, or decoded after the fact from a latent space \citep{chemeleon2rl}. The recent work SPARC \citep{sparc} applies a reward-weighted denoising loss to the discrete atom types of the symmetry-constrained SymmCD generator \citep{symmcd}, but without a likelihood ratio, a group baseline or clipping.

Reward hacking, where a policy scores well on a proxy without achieving the intended goal, is a known failure of reinforcement learning \citep{amodei2016,skalse} that grows with optimization pressure \citep{gao2023overopt} and is documented for diffusion fine-tuning \citep{zhang2024overopt} and KL-regularized RL \citep{catgoodhart,klmodecollapse}. In crystal generation the reported failure is collapse onto a few compositions, met with diversity terms \citep{matinvent,chemeleon2rl}, reweighted uniqueness \citep{csun} or multiplicative aggregation of the objectives \citep{crystalicml2026}. On the evaluation side, the stable, unique and novel (S.U.N.) metric introduced with MatterGen \citep{mattergen} is now standardized by LeMat-GenBench \citep{lematgenbench}, and its limits are under study. Enumeration baselines match generative models on it \citep{szymanski2025baselines}, many metastable generations are duplicates or substitution variants of training structures \citep{negishi2026substitution,martirossyan2025glitter}, and every stability claim inherits the error of the potential behind it \citep{matbenchdiscovery}.

\section{Methods}\label{sec:method}

\subsection{Background: stochastic interpolants and masked discrete flow matching}\label{sec:background}

OMatG uses the unit-cell representation for crystalline materials where a unit cell with $N$ atoms
is described by its atomic numbers $A = ( A^1,...,A^N)$, its fractional coordinates $X = (x^1,...,x^N)$ and its lattice matrix $L$ \citep{omatg}, and we write $s = (A, X, L)$ for this state.
The continuous variables $X$ and $L$ are modeled through the stochastic interpolants framework. In one possible choice in its flexible implementation, OMatG learns a velocity field $b_\theta(t, s_t)$ and a denoiser $z_\theta(t, s_t)$ to interpolate between samples from a base distribution $\rho_0$ at time $0$ and crystal structures following the data distribution $\rho_1$ at time $1$.
During generation, a sample $s_0$ from the base distribution is transported to a sample $s_1$ by integrating, jointly for $X$ and $L$, an SDE with drift $\mu_\theta(s_t) = b_\theta(t, s_t) - \frac{\epsilon(t)}{\gamma(t)}\, z_\theta(t, s_t)$ and noise amplitude $\sqrt{2\epsilon(t)}$, where $\gamma(t)$ was fixed during training and $\epsilon(t)$ can be chosen at inference \citep{omatg,omatgirl}. Euler--Maruyama steps of size $\Delta t$ turn each step of each variable into a Gaussian transition kernel (Appendix~\ref{app:derivation}).

The species, atom types $A$, are modeled through masked discrete flow matching \citep{campbell2024}. Each $A_t^d$ takes a value in $\{M, 1, \dots, S\}$, where $M$ is the mask state and $S$ the number of elements, and OMatG learns the denoising probability $p_\theta(A_1^d = j \mid s_t)$ of the final species of atom $d$ given the current state. During generation all atoms start masked, and the denoising probabilities define the rate matrix $R^\star_t$ of a continuous-time Markov chain that progressively unmasks them, integrated in Euler steps on the same time grid as the SDE, so that every atom carries an element at time $1$. A second rate matrix $R^{\mathrm{DB}}_t$ can add remasking with a noise parameter $\eta$.

\subsection{GRPO on the native composition channel}\label{sec:recipe}

We use GRPO \citep{deepseekmath} as our RL algorithm. A group of $K$ rollouts is drawn
from one initial condition, the same noise and masks, and diverges throughout the stochastic generation dynamics. Each rollout $i$ receives a scalar reward $r^i$ that is computed based on the final structure. Our reward function is defined in Section~\ref{sec:reward}. The
advantage of a rollout is taken relative to its own group,
\begin{equation}
\hat{A}^i \;=\; \frac{r^i - \mathrm{mean}_j\, r^j}{\mathrm{std}_j\, r^j}.
\label{eq:adv}
\end{equation}
The policy is the sampler's transition kernel $\pi_\theta(\cdot \mid s_t)$ over one integration step of size $\Delta t$. Positions, lattice and species are the three channels of the generator, indexed by $f$.
Within a step the sampler first advances positions and lattice from $s_t$ and then draws the species from the updated positions and lattice, so the probability of a step is a product of one factor per channel.
The likelihood ratio between the current and the old policy is written per channel as
\begin{equation}
q^f_t(\theta) \;=\;
\frac{\pi^f_\theta(s^f_{t+\Delta t}\mid s_t)}
     {\pi^f_{\theta_\mathrm{old}}(s^f_{t+\Delta t}\mid s_t)},
\label{eq:ratio}
\end{equation}
and the parameters are updated by maximizing $J(\theta) = \sum_f J^f(\theta)$, the sum over channels of the clipped surrogate objective \citep{ppo2017}
\begin{equation}
J^f(\theta) \;=\; \frac{\alpha_f}{|\mathcal{T}^f|}\sum_{(i,t)\in\mathcal{T}^f}
\min\!\big(q^f_t(\theta)\,\hat{A}^i,\;
\mathrm{clip}(q^f_t(\theta),\,1-\epsilon_{\mathrm{clip}},\,1+\epsilon_{\mathrm{clip}})\,\hat{A}^i\big)
\;-\; \beta_f\, D^f_{\mathrm{KL}}(\theta),
\label{eq:grpo}
\end{equation}
where $\mathcal{T}^f$ collects the update points of channel $f$ across the group. For the positions and lattice, these are the integration steps of every rollout, and for the composition they are the commit events of every atom. Dividing by $|{\mathcal{T}^f}|$ averages each channel over its own update points, and every update point of a trajectory carries the advantage $\hat{A}^i$ of the structure.
The weights $\alpha_f$ compensate for the different scales of the channels, since one position step carries the log-density of all atoms while one commit event carries a single categorical log-probability, and are set so that the gradient magnitudes of the channels are comparable (Appendix~\ref{app:config}).
The term $D^f_{\mathrm{KL}}$ is a KL penalty toward the frozen pretrained model with weight $\beta_f$, as detailed in Appendix~\ref{app:derivation}.

The ratio of Eq.~\ref{eq:ratio} takes a simple form for the composition channel, which we derive next. Each step is a categorical transition that
factorizes over atoms, $\pi^{\mathrm{species}}_\theta(A_{t+\Delta t}\mid s_t) = \prod_d \pi^d_\theta(A^d_{t+\Delta t}\mid s_t)$. In discrete flow matching,
a masked atom keeps its mask with probability $1-\Delta t/(1-t)$ and
otherwise commits to element $j$ with probability $\frac{\Delta t}{1-t}\,p_\theta(A_1^d = j\mid s_t)$. This prediction is the only model-dependent part of the species channel. The factors $1-\Delta t/(1-t)$ and ${\Delta t}/{(1-t)}$ come from the schedule alone. They are the same under the current and the old policy and cancel in the ratio. Let $U_t$ be the set of atoms that commit at step $t$. The ratio for one step is then
\begin{equation}
q^{\mathrm{species}}_t(\theta)
  = \prod_{d\in U_t}
    \frac{p_\theta(A_1^d = A^d_{t+\Delta t}\mid s_t)}
         {p_{\theta_{\mathrm{old}}}(A_1^d = A^d_{t+\Delta t}\mid s_t)}.
\label{eq:qspecies}
\end{equation}
Nothing in Eq.~\ref{eq:qspecies} is approximated, and the policy gradient reaches the
composition channel exactly. Each factor of this product is one update point of Eq.~\ref{eq:grpo}. At the last step the schedule factor equals one, so every atom still masked commits to a valid species and enters $U_T$. We train without species noise, so a committed atom keeps its element. Appendix~\ref{app:derivation} shows that the ratio keeps the form of Eq.~\ref{eq:qspecies} with species noise, where an atom can return to the mask state at a rate that carries no model dependence, and also gives the KL term of Eq.~\ref{eq:grpo} for this channel.
\subsection{Reward stack and guards}\label{sec:reward}
This section describes the reward function in its final form
(Fig.~\ref{fig:concept}a). Removing components of the reward opens specific exploits, as discussed in Section~\ref{sec:hack}. Every final rollout structure is first relaxed inside the reward, with at most 100 FIRE steps including the cell under the UMA potential \citep{uma}, and all terms are computed on the relaxed structure (Appendix~\ref{app:config}). 

The stability term uses the energy above the convex hull, $E_{\mathrm{hull}}$, measured against the UMA-consistent LeMat-Bulk hull released with LeMat-GenBench \citep{lematgenbench}. The reward term is $r_{\mathrm{stab}} = -\mathrm{clip}(\ehull, 0, 1)$, the form used by Chemeleon2 \citep{chemeleon2rl}, and a structure on or below the hull scores zero. Three rules guard the stability term. A structure whose chemical subsystem has fewer than 12 near-hull references abstains, which zeroes its advantage and removes it from the group baseline. We choose this abstention route because a stability claim against so few references is unreliable rather than wrong. A single-element structure receives the reward floor $r_{\mathrm{pen}}$, because it is always sparsely referenced and would otherwise become an unpunished region that group-relative training drifts into (Section~\ref{sec:hack}). A structure with more than 0.1~eV/atom below the hull receives the floor as well, because under the potential that defines the hull such a value marks an unphysical failure of the potential rather than a realistic discovery.

The creativity term, adapted from Chemeleon2 \citep{chemeleon2rl}, rewards a structure for being unique and novel. A relaxed structure is novel if no structure of the same reduced formula in the MP-20 training set \citep{cdvae, materialsproject} matches it under the pymatgen structure matcher \citep{pymatgen}, and unique if no earlier structure of the same formula in the rollout batch matches it structurally. A structure that is both scores 1, one that is neither scores 0, and the mixed cases receive partial credit defined in Appendix~\ref{app:config}. Two diversity terms discourage repetition at two scales. Within each rollout group, a composition-occurrence discount $c_{\mathrm{occ}}\in[0,1]$, adapted from MatInvent \citep{matinvent}, pulls the reward of a repeated composition toward the floor $r_{\mathrm{pen}}$, because the group would otherwise reward local duplication. Across the entire batch consisting of several rollout groups, a leave-one-out maximum-mean-discrepancy bonus \citep[MMD;][]{gretton2012mmd} credits each structure by how much its composition reduces the discrepancy between the batch and the MP-20 composition distribution, because stability-led rewards could otherwise narrow the chemistry of the whole batch in a mode collapse. In summary, the combined reward is
\begin{equation}
r \;=\; r_{\mathrm{pen}} \;+\; c_{\mathrm{occ}}\,\bigl[\, r_{\mathrm{stab}} + r_{\mathrm{creat}} - r_{\mathrm{pen}} \,\bigr] \;+\; 0.4\, r_{\mathrm{mmd}},
\label{eq:reward}
\end{equation}
with $r_{\mathrm{pen}}=-1$. The kernel and per-batch rescaling of the MMD term, the partial-credit rule, and the routing logic are given in Appendix~\ref{app:config}.

\section{Experiments}\label{sec:experiments}

\begin{table}[t]
\caption{Yield of mSUN under LeMat-GenBench. Valid is the share of the nominal 2,500 generations that passes the benchmark's structural checks. Unique, novel and metastable are rates over valid structures, the benchmark's own convention, and mSUN is reported over the nominal 2,500 so that methods are comparable (counts in parentheses). SUN and single-element mSUN are counts. Prior is the pretrained OMatG model. The zero single-element and sparse-hull entries of best-of-$N$ are enforced by its selection score rather than learned, and the controls are defined in Section~\ref{sec:gains}.}
\label{tab:master}
\centering
\footnotesize
\setlength{\tabcolsep}{3pt}
\begin{tabular}{@{}lcccccccc@{}}
\toprule
method & valid & unique & novel & metastable & SUN & \textbf{\msun{}} & single-element & sparse hull \\
 & \% $\uparrow$ & \% $\uparrow$ & \% $\uparrow$ & \% $\uparrow$ & count $\uparrow$ & \% (count) $\uparrow$ & in \msun{} $\downarrow$ & \% of \msun{} $\downarrow$ \\
\midrule
prior                    & 94.1 & 99.5 & 82.1 & 27.2 & 8  & 13.4 (336) & 7 & 21.7 \\
best-of-$N$              & 97.9 & 77.9 & 11.5 & 65.6 & 72 & 7.7 (192)  & 0 & 0.0 \\
frozen control           & 93.5 & 99.5 & 87.7 & 18.7 & 9  & 9.8 (245)  & 1 & 12.2 \\
Chemeleon2               & 94.9 & 90.6 & 68.1 & 67.5 & 36 & 36.6 (915) & 9 & 7.9 \\
\textbf{OMatGRPO (ours)} & 69.8 & 98.7 & 88.3 & 75.8 & 7  & \textbf{45.5 (1138)} & 4 & 23.5 \\
\bottomrule
\end{tabular}
\end{table}

\subsection{Setup}\label{sec:setup}
Training follows the GRPO algorithm defined in Section~\ref{sec:recipe}. Each rollout iteration draws 4 groups of 16 replicates with the training-time sampler (64 steps). All RL training runs start from an OMatG model pretrained on MP-20. Generated structures are scored with the reward function defined by Eq.~\ref{eq:reward}. Each rollout batch is reused for three inner policy updates, and training runs for 750 rollout iterations. Full hyperparameters are in
Appendix~\ref{app:config}.


Every score in Tables~\ref{tab:master} and~\ref{tab:swap} comes from LeMat-GenBench~\citep{lematgenbench}, run unmodified. \msun{} is the fraction of generated structures that are metastable, unique, and novel at once. Stability is scored by a three-MLIP ensemble
(Orb, MACE, UMA) with agreement from at least two required and a metastability threshold of 0.1~eV/atom, and uniqueness and novelty from the benchmark's structure matcher against the roughly 5.3 million LeMat-Bulk structures. Each method submits a nominal 2{,}500 structures, and every \msun{} rate in the paper is a fraction of 2{,}500, so methods are comparable regardless of how
many structures survive the benchmark's validity checks.


\subsection{RL triples the yield of stable, new compounds}\label{sec:gains}

\begin{figure}[t]
\centering
\includegraphics[width=\linewidth]{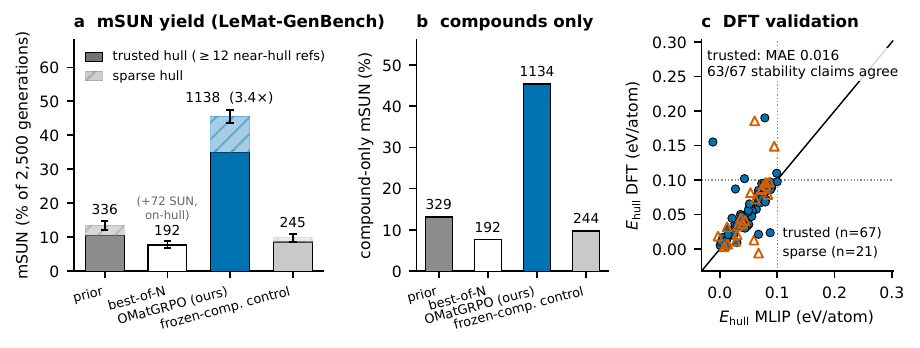}
\caption{\textbf{Reinforcement learning gains, their controls, and DFT validation.} Every method submits 2,500 generated structures, relaxed and scored by LeMat-GenBench (Section~\ref{sec:setup}). Counts are printed above the bars. \textbf{(a)} Solid segments sit on trusted hulls with at least 12 near-hull references, lighter segments of the same color on sparse hulls (Section~\ref{sec:reward}), and error bars are 95\% confidence intervals. The controls are defined in Section~\ref{sec:gains}. \textbf{(b)} Compounds only, which removes the single-element structures of Section~\ref{sec:hack}. \textbf{(c)} MLIP against DFT stability claims for the spot-checked structures. On trusted hulls ($n{=}67$) the mean absolute deviation is 0.016 eV/atom and 63 of 67 claims agree. On sparse hulls ($n{=}21$) the energies track but stability cannot be adjudicated. Appendix~\ref{app:dft} lists every structure, the four disagreements and the two excluded cases.}
\label{fig:gains}
\end{figure}

The first question is whether RL improves the yield of stable, new compounds at all. After 750 rollout iterations, \msun{} rises from the prior's \priorrate{} (336
of 2{,}500) to \headlinerate{} (1{,}138 of 2{,}500), a \gainx{} gain
(Fig.~\ref{fig:gains}a, Table~\ref{tab:master}). Chemeleon2, the
published work we compare against, reaches 36.6\% (915 of 2{,}500) under the
identical benchmark run.

Two control methods, frozen-composition channel and best-of-$N$, test two things. The frozen-composition control measures the contribution of species RL by turning it off, and best-of-$N$ tests whether RL's gain can be achieved with oversampling. Best-of-$N$ selection, which keeps the top 2{,}500 of 47{,}795 prior samples ranked by the stability reward, reaches an \msun{} of 7.7\%, a sixth of the RL-trained policy's rate. What it surfaces is mostly
rediscovery, with a novelty rate of 11.5\%. The frozen-composition control reaches 9.8\%, below the prior itself. Under the same rollout budget, RL restricted to the position and lattice channels with the stability term alone does not reproduce the gain (Appendix~\ref{app:config}), and its \msun{} set (244 of its 245 structures are compounds) stays close to
the prior's composition profile. For a detailed study of RL restricted to the position and lattice channels of OMatG we refer to OMatG-IRL \citep{omatgirl}. This suggests that most of the full OMatGRPO gain in Table~\ref{tab:master} comes from the discrete species RL. 
Fig.~\ref{fig:gains}b restricts the comparison to compounds. The OMatGRPO \msun{} set contains 1{,}134 compounds and 4 single-element structures (0.4\% of the set), so it does not suffer from the single-element exploit that Section~\ref{sec:hack} discusses.

Every stability number to this point, in the reward and in the benchmark, comes from MLIPs, and a policy could raise its benchmark score by drifting toward compositions where the potentials err, so we check the metastability claims with DFT. Across the spot-checks accumulated in this work, 63 of 67 trusted-tier metastability
claims are confirmed, with a mean absolute deviation of 0.016~eV/atom between
UMA and DFT hull distances (Fig.~\ref{fig:gains}c). For the OMatGRPO model alone the count is 20 of 22. Appendix~\ref{app:dft} gives the DFT protocol and the per-structure results.

\subsection{Reward hacking}\label{sec:hack}

\begin{figure}[t]
\centering
\includegraphics[width=\linewidth]{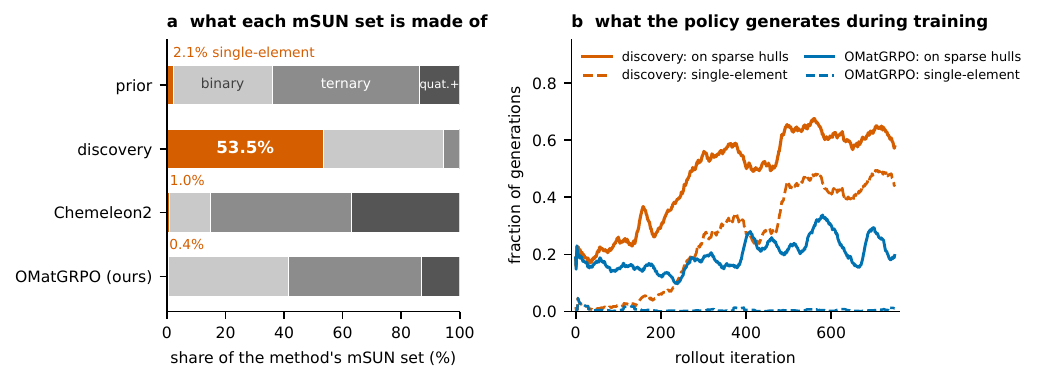}
\caption{\textbf{Reward hacking via single-element structures.} Throughout, \emph{discovery} is the reward of
Eq.~\ref{eq:reward} with the single-element guard removed. \textbf{a}, Composition of each method's \msun{}
set, grouped by the number of distinct elements per structure, with the pretrained prior and Chemeleon2 for reference.
\textbf{b}, Fraction of generations scored on sparse hulls and fraction that are single-element during training, smoothed over 25 rollout iterations, for the discovery run and OMatGRPO.}
\label{fig:hack}
\end{figure}

Each term and guard of Eq.~\ref{eq:reward} was added in response to a failure of
the reward that preceded it, where the policy found a way to score well
without producing better materials. We describe that sequence here because it
shows which parts of the reward are necessary and why, and Appendix~\ref{app:hacks}
catalogs every variant with its per-run numbers.


Without the clip on $E_\mathrm{hull}$ in Eq.~\ref{eq:reward}, the reward grows without limit as a structure falls below the hull, and within a group the highest reward goes to the lowest $E_\mathrm{hull}$. The policy migrated into a narrow family of Li--O compositions that the MLIP places far below its own convex hull, an artifact of the energy model and of the few competing phases there rather than of true stability.
Clipping the reward at the hull removed this incentive, but the clipped run drifted instead onto thinly referenced hulls, where
$E_\mathrm{hull}\!\approx\!0$ is easy to reach because the reference contains
little to compete against (Section~\ref{sec:soft}). We responded by withholding reward on sparse hulls, so that a sparsely referenced $E_\mathrm{hull}$ earns abstention instead of
credit.


Abstention created the next exploit, because a structure on a sparse hull is never punished, and the policy drifted into that region. In the ablation with the same reward without the single-element guard (the discovery run of Fig.~\ref{fig:hack}), the fraction of generations routed to sparse hulls climbed from 0.19 to 0.58 over training (Fig.~\ref{fig:hack}b), and the growth was almost entirely single-element structures.
A repacked cell of a common metal passes every \msun{} filter for the reasons given in Section~\ref{sec:intro}, and fcc Mg in Fig.~\ref{fig:concept} is the clearest example. Under LeMat-GenBench 53.5\% of the discovery run's \msun{} set is single-element (Fig.~\ref{fig:hack}a), and MatterGen's evaluation code and Chemeleon2's \msun{} definition accept these structures as well.

The exploit is induced by the training. The MP-20 prior already emits a few single-element cells, but a second prior trained on Alexandria, roughly twenty times larger, places none in its \msun{} set (0 in 1{,}000) and 104 per 1{,}000 after RL (Appendix~\ref{app:hacks}). We close the exploit by assigning every single-element generation the penalty reward $r_\mathrm{pen}$ of Eq.~\ref{eq:reward}. With the guard, single-element structures make up 0.4\% of the OMatGRPO \msun{} set (Table~\ref{tab:master}), the guard holds across three seeds (Appendix~\ref{app:hacks}), and during training the elemental fraction remains near zero while the sparse-routed fraction holds near 0.2 (Fig.~\ref{fig:hack}b).


Our reward and the benchmark share components, so the gain has to be read with that circularity in mind. Novelty and uniqueness are two of the three components of \msun{} and our creativity term rewards both. Without the creativity term the reward reaches 39.3\% \msun{} against the prior's 13.4\% (Appendix~\ref{app:internal}), so about 80\% of the gain comes from a reward that optimizes neither. We also note that novelty was rewarded against the MP-20 training set, while the benchmark tests it against roughly five million LeMat-Bulk structures that include essentially all of MP-20, a harder test. Stability is scored by one potential in the reward and by an ensemble of three in the benchmark, of which UMA is one, against hulls built from the same reference data, and DFT is the one check no reward used (Section~\ref{sec:soft}).

\subsection{Swapping rewards across generators}\label{sec:swap}

\begin{table}[t]
\caption{Swapping rewards between the two pipelines. Each cell is a policy trained with the row's reward inside the column's pipeline and evaluated as in Table~\ref{tab:master}, given as \msun{} in \% of the nominal 2{,}500 with counts in parentheses. The diagonal holds OMatGRPO and the released Chemeleon2 checkpoint. Each cell keeps its pipeline's relaxation convention, ours relaxes before the reward and Chemeleon2's does not, and Appendix~\ref{app:swap} reports both variants.}
\label{tab:swap}
\centering
\small
\setlength{\tabcolsep}{6pt}
\begin{tabular}{lcc}
\toprule
 & \multicolumn{2}{c}{trained in pipeline} \\
\cmidrule(lr){2-3}
reward & \textbf{ours} (native-channel SDE) & Chemeleon2 (latent VAE) \\
\midrule
\textbf{ours}  & \textbf{45.5} (1138) & 43.7 (1093) \\
Chemeleon2       & 18.6 (465)           & 36.6 (915) \\
\bottomrule
\end{tabular}
\end{table}

To separate what the reward contributes from what the training pipeline
contributes, we exchanged rewards between OMatGRPO and Chemeleon2
\citep{chemeleon2rl} and trained a policy for each. The results are the off-diagonal cells of
Table~\ref{tab:swap}. Each reward was reimplemented inside the other pipeline with its terms and weights unchanged, and every run was evaluated by the protocol of Table~\ref{tab:master}. OMatGRPO relaxes each structure before computing the reward, while Chemeleon2's reward scores the decoded structure directly.

Our reward produces the higher \msun{} in both pipelines (Table~\ref{tab:swap}). In our pipeline it reaches 45.5\% against 18.6\% for Chemeleon2's reward, and in Chemeleon2's pipeline it reaches 43.7\% against the 36.6\% of the released model trained on its native reward. All four cells sit above the prior's 13.4\%.
Under Chemeleon2's reward our pipeline also leaks single-element structures into its \msun{} set, 4.1\% with relaxation before scoring and 7.5\% without, as no term in their reward accounts for the number of elements. Their own pipeline rarely meets the problem, since its frozen decoder trained on MP-20 seldom decodes a single-element cell (9 of 915 \msun{} structures in Table~\ref{tab:master}), while our generator writes the composition directly. Appendix~\ref{app:swap} interprets the two off-diagonal cells, marked as interpretation, with the supporting diagnostics and the open checks.


\subsection{A stability claim is only as good as its reference hull}\label{sec:soft}

The metastability inside mSUN is a relative measure, because $E_\mathrm{hull}$ is a distance to the convex hull of known competing phases. On a hull with only a few references a value near zero says only that the structure is no worse than the few phases it was compared against, and every benchmark that scores stability against a reference set inherits this limit. In the pretrained prior, 21.7\% of the mSUN set sits on hulls with fewer than 12 reference phases, while the discovery run of Section~\ref{sec:hack} has 63.4\% of its mSUN set there (Table~\ref{tab:master} and Appendix~\ref{app:internal}), because RL moves the generator toward the regions where a metastable label is easiest to obtain. OMatGRPO holds the share at 23.5\% (Fig.~\ref{fig:gains}a), close to the prior's level.

The limiting factor is the reference hull rather than the potential. A second machine-learned potential reproduces the stability energies to a median of 0.015 eV/atom, and on the 21 spot-checked structures that fall on sparse hulls DFT tracks UMA as well as on the trusted tier yet cannot adjudicate the claim, because the DFT hull is missing the same phases (Appendix~\ref{app:dft}). Sparse-tier mSUN is therefore reported in every result.



\section{Conclusion}\label{sec:conclusion}
We reinforced the composition channel of a crystal generator directly. Under masked discrete flow matching every atom type is a commit event with an exact likelihood, so GRPO updates the elements of a structure as actions of the policy. The reward relaxes each structure, scores its distance to the convex hull, its novelty and uniqueness, and the diversity of the batch, and withholds credit where the reference hull is sparse. With it, the yield of metastable, unique and novel structures under LeMat-GenBench rises from 13.4\% to 45.5\%. DFT confirms 63 of 67 spot-checked stability claims, and the same reward raises Chemeleon2 from 36.6\% to 43.7\% inside its own pipeline. Each guard in the reward closed an exploit the policy had found, and the last of them, the repacking of single elements into new cells, is counted as discovery by every mSUN implementation we checked. A stability claim is only as good as its reference hull, so we report the number of reference phases behind every structure and read sparse-tier mSUN as candidates rather than discoveries. Benchmarks already compute these counts and should report them, and a promising sparse candidate can be settled by computing its missing competing phases with DFT.

\subsubsection*{Reproducibility statement}
The policy objective and the composition-channel ratio are given in Section~\ref{sec:recipe}, and Appendix~\ref{app:derivation} derives the ratio and the KL terms in the form the implementation evaluates them. The reward is defined in Section~\ref{sec:reward} and Eq.~\ref{eq:reward}, and Appendix~\ref{app:config} lists its thresholds, the best-of-$N$ selection rule and every training hyperparameter. All benchmark numbers come from LeMat-GenBench run unmodified on 2{,}500 generations per method (Section~\ref{sec:setup}), and Appendix~\ref{app:internal} maps the internal evaluation of Section~\ref{sec:hack} onto that convention. The reward-swap protocol and its recorded deviations are in Appendix~\ref{app:swap}, and the DFT protocol and per-structure results in Appendix~\ref{app:dft}. Code and run configurations are available at \url{https://github.com/paprakash/OMatGRPO}, and the trained models and generated structure sets at \url{https://huggingface.co/paprakash/OMatGRPO}.

\subsubsection*{Ethics statement}
This work uses only public data (MP-20, Alexandria and the LeMat-Bulk reference set) and involves no human subjects or personal information. The reward targets thermodynamic stability, novelty and diversity of inorganic crystals and no property tied to a harmful use, so we do not foresee risks specific to this work beyond those of computational materials discovery in general. Because the paper shows that a reinforced policy can raise a community benchmark without producing new materials, we report every result split by the number of reference phases behind it, so that readers can judge which stability claims are supported.

\subsubsection*{LLM usage disclosure}
We used Claude (Anthropic) in this work.
It helped implement and review the training and analysis code, run experiments and make figures. It searched and summarized the related literature, and helped draft the derivation in Appendix~\ref{app:derivation}, which we verified ourselves. It also drafted parts of the text, which the authors rewrote or edited. It was not used to generate data. The authors reviewed all AI-assisted work and take responsibility for the final content of this work, including text, claims and artifacts produced with its aid.

\bibliography{iclr2027_conference}
\bibliographystyle{iclr2027_conference}

\appendix
\input{appendix_derivation}
\input{appendix_internal}
\input{appendix_b_hack}
\input{appendix_c}
\input{appendix_dft}
\input{appendix_config}

\end{document}

%% file: math_commands.tex
\usepackage{amsmath,amsfonts,bm}

\def\eqref#1{equation~\ref{#1}}

\def\1{\bm{1}}

\DeclareMathAlphabet{\mathsfit}{\encodingdefault}{\sfdefault}{m}{sl}
\SetMathAlphabet{\mathsfit}{bold}{\encodingdefault}{\sfdefault}{bx}{n}



%% file: appendix_derivation.tex

\section{Derivation of the policy ratio and the KL terms}\label{app:derivation}

This appendix derives the composition-channel ratio of Eq.~\ref{eq:qspecies} for any
species noise $\eta \geq 0$ and gives the KL terms of Eq.~\ref{eq:grpo} in the form
the implementation evaluates them. It rests on the continuous-time Markov chain
construction of masked discrete flow matching by \citet{campbell2024}, the species
model of OMatG \citep{omatg}. We use the notation of Section~\ref{sec:recipe}. A
structure has $N$ atoms, $A_t^d \in \{M, 1, \dots, S\}$ is the species of atom $d$ at
time $t$ with $M$ the mask state, generation starts fully masked and ends with no atom
masked, and the sampler takes $T$ Euler steps of size $\Delta t$. The model output
$p_\theta(A_1^d = j \mid s_t)$, abbreviated $p^d_\theta(\cdot \mid s_t)$, is the
predicted final species of atom $d$. For the composition channel $s_t$ holds the
species at time $t$ and the positions and lattice already advanced to $t + \Delta t$,
since the sampler moves the continuous channels first within a step. The reference
policy $\pi_{\mathrm{ref}}$ is the frozen pretrained checkpoint with predictions
$p_{\mathrm{ref}}$, run with the same $\eta$ and the same time grid.

\subsection{Masked DFM as a continuous-time Markov chain}

Masked DFM connects the fully masked state to the data through the conditional
interpolant
\begin{equation}
p_{t\mid 1}(A_t \mid A_1) \;=\; \prod_{d=1}^{N}
\big[\, t\,\delta_{A_t^d,\,A_1^d} + (1-t)\,\delta_{A_t^d,\,M} \,\big],
\label{eq:app-interp}
\end{equation}
so at time $t$ each atom independently shows its final species with probability $t$
and the mask otherwise \citep[Eq.~6]{campbell2024}. Generation simulates a Markov chain
with rate matrix $R^\star_t + \eta R^{\mathrm{DB}}_t$, which generates this interpolant
for every $\eta \geq 0$ \citep[Prop.~3.3 and App.~F.1]{campbell2024}, with the unknown
final species replaced by the model's prediction. For atom $d$ the only nonzero
off-diagonal rates are
\begin{equation}
R^{d}_t(M, j) \;=\; \frac{1+\eta t}{1-t}\;p_\theta(A_1^d = j \mid s_t)
\quad\text{for } j \neq M,
\qquad
R^{d}_t(k, M) \;=\; \eta \quad\text{for } k \neq M .
\label{eq:app-rates}
\end{equation}
The model enters only through the unmask rate. The remask rate is a constant of the
sampler, and there is no direct move between two species.

\subsection{The Euler step kernel}

Over one step every atom moves independently given $s_t$, so the step kernel
factorizes over atoms, $\pi^{\mathrm{species}}_\theta(A_{t+\Delta t}\mid s_t) =
\prod_{d} \pi^d_\theta(A^d_{t+\Delta t}\mid s_t)$. With
$p_{\mathrm{unmask}}(t) = \min\{1, \frac{1+\eta t}{1-t}\,\Delta t\}$ and
$p_{\mathrm{remask}} = \eta\,\Delta t$ the per-atom kernel is
\begin{equation}
\pi^d_\theta(A^d_{t+\Delta t}\mid s_t) \;=\;
\begin{cases}
1 - p_{\mathrm{unmask}}(t)
  & \text{if } A_t^d = M,\ A^d_{t+\Delta t} = M,\\[2pt]
p_{\mathrm{unmask}}(t)\,p_\theta(A_1^d = j \mid s_t)
  & \text{if } A_t^d = M,\ A^d_{t+\Delta t} = j \neq M,\\[2pt]
p_{\mathrm{remask}}
  & \text{if } A_t^d = k \neq M,\ A^d_{t+\Delta t} = M,\\[2pt]
1 - p_{\mathrm{remask}}
  & \text{if } A_t^d = k \neq M,\ A^d_{t+\Delta t} = k,\\[2pt]
0 & \text{otherwise.}
\end{cases}
\label{eq:app-kernel}
\end{equation}
The implementation draws the unmask and remask decisions as Bernoulli variables with
these probabilities and the species of an unmasking atom from
$p^d_\theta(\cdot \mid s_t)$. At $\eta = 0$ an unmasked atom keeps its species with
probability one, the setting of every run in this paper. Writing
$I^d_{\mathrm{um}} = \mathbf{1}[A_t^d = M,\ A^d_{t+\Delta t} \neq M]$ for an unmask
transition, the log kernel is
\begin{equation}
\log \pi^d_\theta(A^d_{t+\Delta t}\mid s_t)
  = \log c^d_t \;+\; I^d_{\mathrm{um}}\,
     \log p_\theta\big(A_1^d = A^d_{t+\Delta t} \mid s_t\big),
\label{eq:app-logkernel}
\end{equation}
where $\log c^d_t$ collects the schedule factors of the four cases and depends only on
$t$, $\eta$ and $\Delta t$. The model appears in a single term, and only at atoms that
unmask in this step. Transitions outside the four cases have probability zero under
$\pi_\theta$, $\pi_{\theta_{\mathrm{old}}}$ and $\pi_{\mathrm{ref}}$ alike, so every
quantity below is evaluated on their common support.

\subsection{The policy ratio}

The composition-channel ratio of Eq.~\ref{eq:ratio} is the product over atoms of
$\pi^d_\theta / \pi^d_{\theta_{\mathrm{old}}}$. By Eq.~\ref{eq:app-logkernel} the
factors $c^d_t$ cancel and every atom that does not unmask contributes a factor of one.
With $U_t$ the set of atoms that commit at step $t$,
\begin{equation}
q^{\mathrm{species}}_t(\theta)
  \;=\; \prod_{d \in U_t}
    \frac{p_\theta\big(A_1^d = A^d_{t+\Delta t} \mid s_t\big)}
         {p_{\theta_{\mathrm{old}}}\big(A_1^d = A^d_{t+\Delta t} \mid s_t\big)},
\label{eq:app-ratio}
\end{equation}
which is Eq.~\ref{eq:qspecies}. Nothing in this step used $\eta = 0$. Remasking adds
transitions with $\theta$-independent factors that drop out in the same way. What
changes with $\eta$ is the set of update points. At $\eta = 0$ every atom commits
exactly once and a rollout with $N$ atoms contributes $N$ update points to
$\mathcal{T}^{\mathrm{species}}$. With $\eta > 0$ a remasked atom commits again later
and each commit is a further update point, evaluated at the state of its own step. The
surrogate of Eq.~\ref{eq:grpo} clips every factor of Eq.~\ref{eq:app-ratio}
separately. At the last integration step the implementation unmasks every atom that is
still masked and draws no remask, so $p_{\mathrm{unmask}} = 1$, the forced commits
enter $U_t$ with weight one, and Eq.~\ref{eq:app-ratio} keeps its form.

\subsection{KL term for the composition channel}

Both step kernels factorize over atoms, so the KL divergence between the current and
the reference kernel is a sum of per-atom divergences. For an unmasked atom the kernel
does not depend on $\theta$ and the divergence vanishes. A masked atom keeps its mask
with the same probability under both policies or unmasks to species $j$ with
probability $p_{\mathrm{unmask}}(t)$ times the respective prediction, so the shared
factor cancels inside the logarithm and
\begin{equation}
D_{\mathrm{KL}}\big[\pi^d_\theta(\cdot\mid s_t)\,\big\|\,
                    \pi^d_{\mathrm{ref}}(\cdot\mid s_t)\big]
\;=\; \mathbf{1}[A_t^d = M]\;p_{\mathrm{unmask}}(t)\;
D_{\mathrm{KL}}\big[p^d_\theta(\cdot\mid s_t)\,\big\|\,
                    p^d_{\mathrm{ref}}(\cdot\mid s_t)\big],
\label{eq:app-klatom}
\end{equation}
with the ordinary categorical KL between the two species predictions on the right.
This holds for any $\eta \geq 0$. The weight $p_{\mathrm{unmask}}(t)$ grows toward the
end of generation and equals one at the last step. Writing $\mathcal{B}$ for the
rollouts of a batch and $N_i$ for the atoms of rollout $i$, the term entering
Eq.~\ref{eq:grpo} is this divergence summed over the masked atoms of every step and
averaged per atom, per step and over the batch,
\begin{equation}
D^{\mathrm{species}}_{\mathrm{KL}}(\theta)
\;=\; \frac{1}{|\mathcal{B}|}\sum_{i\in\mathcal{B}}
      \frac{1}{T N_i}\sum_{t}\;\sum_{d:\,A^{i,d}_t = M}
      p_{\mathrm{unmask}}(t)\;
      D_{\mathrm{KL}}\big[p^d_\theta(\cdot\mid s^i_t)\,\big\|\,
                          p^d_{\mathrm{ref}}(\cdot\mid s^i_t)\big].
\label{eq:app-klspecies}
\end{equation}
It is evaluated at every masked atom of every step, whether or not the atom commits,
so no Bernoulli draw enters it. The implementation evaluates it for $\eta = 0$, where
$p_{\mathrm{unmask}}(t) = \min\{1, \Delta t/(1-t)\}$, with
$\beta_{\mathrm{species}} = 0.05$ (Appendix~\ref{app:config}).

\subsection{Continuous channels}

Positions and lattice follow an Euler--Maruyama discretization of the OMatG SDE
sampler with drift $\mu_\theta(s_t)$ and noise schedule $\epsilon(t)$
\citep{omatg, omatgirl}. One step of the positions is
$x^d_{t+\Delta t} = x^d_t + \mu^d_\theta(s_t)\,\Delta t + \sqrt{2\epsilon(t) \Delta t}\,\xi^d$
with $\xi^d$ standard normal, so the step kernel of a structure's positions is a
Gaussian with log-density
\begin{equation}
\log \pi^{\mathrm{pos}}_\theta(X_{t+\Delta t}\mid s_t)
\;=\; -\sum_{d=1}^{N}
  \frac{\big\lVert x^d_{t+\Delta t} - x^d_t - \mu^d_\theta(s_t)\,\Delta t\big\rVert^2}
       {4\,\epsilon(t)\,\Delta t} \;+\; \mathrm{const},
\label{eq:app-poslogp}
\end{equation}
where the displacement is taken to the nearest periodic image and the constant does
not depend on $\theta$. The implementation floors the variance $2\epsilon(t)\Delta t$
at $10^{-4}$. The ratio $q^{\mathrm{pos}}_t(\theta)$ of Eq.~\ref{eq:ratio} is the
ratio of two such densities of the realized step, one per structure and step. The
lattice channel has the same form with the nine lattice components in place of the
atoms and its own noise schedule. The reference policy shares the noise schedule, so
the KL divergence between the two position kernels is a squared difference of drifts,
and averaged as in Eq.~\ref{eq:app-klspecies} the term entering Eq.~\ref{eq:grpo} is
\begin{equation}
D^{\mathrm{pos}}_{\mathrm{KL}}(\theta)
\;=\; \frac{1}{|\mathcal{B}|}\sum_{i\in\mathcal{B}}
      \frac{1}{T N_i}\sum_{t}\sum_{d=1}^{N_i}
      \frac{\big\lVert \mu^d_\theta(s^i_t) - \mu^d_{\mathrm{ref}}(s^i_t)\big\rVert^2\,\Delta t}
           {4\,\epsilon(t)},
\label{eq:app-klpos}
\end{equation}
with $\beta_{\mathrm{pos}} = 0.01$. No KL term acts on the lattice channel, so
$\beta_{\mathrm{cell}} = 0$ and only the clipped surrogate of Eq.~\ref{eq:grpo} acts
on it.

%% file: appendix_internal.tex

\section{Internal evaluation and convention map}\label{app:internal}

Every number in the main text comes from LeMat-GenBench \citep{lematgenbench}. During
development we adjudicated runs with a faster evaluation of our own, and
Section~\ref{sec:hack} and Appendix~\ref{app:hacks} quote counts per $1{,}000$
generations from it. This appendix defines that evaluation, sets it against the
benchmark, and lists the benchmark results of every run the paper names.

\paragraph{The internal evaluation.} For one checkpoint, $1{,}000$ structures are
generated de novo with the training-time sampler, taking only the number of atoms
from a fixed set of MP-20 validation inputs, with the noise seeded identically for
every checkpoint. Each structure is relaxed with FIRE under UMA, including the cell,
to a force tolerance of 0.05~eV/\AA{} within at most 500 steps, and its $\ehull$ is
measured against the UMA-consistent LeMat-Bulk hull together with the number of
reference phases within 1~meV/atom of the hull in its chemical subsystem. A structure
is metastable if $\ehull \leq 0.1$~eV/atom, unique if it is the first representative
of its structure-match cluster within the run, and novel if no structure with the
same reduced formula in the MP-20 training split matches it under the pymatgen
structure matcher with default tolerances, the convention of MatterGen
\citep{mattergen}. \msun{} counts the structures that are all three, over all
$1{,}000$ generations. For the model pretrained on Alexandria, novelty is measured
against the Alex-MP-20 set of \citet{mattergen}, training and validation splits
together, about $600{,}000$ structures.

\paragraph{Convention map.} Table~\ref{tab:conventions} sets the two conventions
side by side. The benchmark is run unmodified at commit \texttt{58e6eae3} with the
preset \texttt{comprehensive\_multi\_mlip\_hull}. Counts under the two conventions are
not interchangeable, and every count the main text quotes from the internal evaluation
has a benchmark counterpart in Table~\ref{tab:allruns} or Table~\ref{tab:master}. The
sparse-hull share of a benchmark \msun{} set is obtained by joining the benchmark's
per-structure \msun{} list with the internal reference count of the same structure.
The join was checked structure by structure against the benchmark's own per-structure
stability values for every run, with no disagreement.

\begin{table}[h]
\caption{The internal evaluation and LeMat-GenBench side by side.}
\label{tab:conventions}
\begin{center}
\small
\begin{tabular}{p{2.8cm}p{4.6cm}p{4.8cm}}
\toprule
 & internal evaluation & LeMat-GenBench \\
\midrule
sample per method & $1{,}000$ generations & $2{,}500$ generations \\
rate basis & count per $1{,}000$ generations & fraction of the nominal $2{,}500$ (\msun) or of the valid structures (other rates) \\
relaxation before scoring & FIRE under UMA with cell, at most 500 steps & our relaxed geometry, then the benchmark's own 50-step relaxation \\
stability model & UMA & Orb, MACE and UMA, each against its own hull, two of three must agree \\
reference hull & UMA-consistent LeMat-Bulk hull & the benchmark's per-potential hulls \\
metastability threshold & $\ehull \leq 0.1$~eV/atom & $\ehull \leq 0.1$~eV/atom \\
reference count & phases within 1~meV/atom of the hull in the chemical subsystem & not reported, joined from the internal evaluation \\
uniqueness and novelty & pymatgen structure matcher, MP-20 training split & the benchmark's structure-matcher fingerprint, LeMat-Bulk (5.3 million structures) \\
\bottomrule
\end{tabular}
\end{center}
\end{table}

\paragraph{Benchmark results of every named run.} Table~\ref{tab:allruns} lists the
LeMat-GenBench results of every run the paper names. The pre-creativity runs are the
development runs before the creativity term was added and differ from their
counterparts only by that term. The penalty-routing runs route every sparsely
referenced structure to the floor instead of to abstention, which removes the
single-element exploit as well, at a lower yield than the guard.

\begin{table}[h]
\caption{LeMat-GenBench results of every run named in the paper, on the nominal
$2{,}500$ basis. Single-element and sparse-hull entries are shares of the run's \msun{}
set, and compounds is the \msun{} count without the single-element structures. The
identifier is the run name in the released code and data, without the suffix
\texttt{\_r750}. The five runs of Table~\ref{tab:master} are repeated for reference.}
\label{tab:allruns}
\begin{center}
\scriptsize
\setlength{\tabcolsep}{3pt}
\begin{tabular}{llrrrrr}
\toprule
run & identifier & \msun{} & \% & single-element & sparse hull & compounds \\
\midrule
prior & \texttt{R0} & 336 & 13.4 & 7 (2.1\%) & 21.7\% & 329 \\
best-of-$N$ & \texttt{bestofn48k\_top2500} & 192 & 7.7 & 0 (0.0\%) & 0.0\% & 192 \\
frozen-composition control & \texttt{frozen\_control} & 245 & 9.8 & 1 (0.4\%) & 12.2\% & 244 \\
Chemeleon2, released model & \texttt{chemeleon2\_mp20rl} & 915 & 36.6 & 9 (1.0\%) & 7.9\% & 906 \\
\midrule
discovery, pre-creativity & \texttt{canonical} & 1063 & 42.5 & 528 (49.7\%) & 65.8\% & 535 \\
discovery (Section~\ref{sec:hack}) & \texttt{canonical\_creatrelax} & 939 & 37.6 & 502 (53.5\%) & 63.4\% & 437 \\
penalty routing, pre-creativity & \texttt{sparseworst} & 789 & 31.6 & 1 (0.1\%) & 0.9\% & 788 \\
penalty routing & \texttt{sparseworst\_creatrelax} & 841 & 33.6 & 1 (0.1\%) & 3.6\% & 840 \\
guarded, pre-creativity & \texttt{arityguard} & 982 & 39.3 & 20 (2.0\%) & 14.1\% & 962 \\
OMatGRPO & \texttt{arityguard\_creatrelax} & 1138 & 45.5 & 4 (0.4\%) & 23.5\% & 1134 \\
\bottomrule
\end{tabular}
\end{center}
\end{table}

\paragraph{The creativity ablation.} The guarded pre-creativity run bounds how much the
overlap between the creativity term and the benchmark matters, as discussed in
Section~\ref{sec:hack}. Without the creativity term the reward reaches 39.3\% \msun{}
against the prior's 13.4\%, and adding the term raises it to 45.5\%, so about 80\% of
the gain comes from a reward that optimizes neither novelty nor uniqueness. Of the two,
uniqueness is the same measurement on both sides, duplicate suppression within a set of
generations, and we do not treat the uniqueness column of Table~\ref{tab:master} as
independent evidence. Novelty was rewarded against the MP-20 training split and is tested
by the benchmark against LeMat-Bulk, which includes essentially all of MP-20.

%% file: appendix_b_hack.tex

\section{Reward-hacking catalog with per-run numbers}\label{app:hacks}

Five reward designs produced five distinct exploits before the reward of
Eq.~\ref{eq:reward} held. Table~\ref{tab:taxonomy} lists them, and the paragraphs
below give the numbers behind each row. Modes 1 to 4 arose during development and are
documented from training-time logs and targeted closure runs. Mode 5 and the
final-recipe runs are adjudicated by the internal evaluation of
Appendix~\ref{app:internal}, and counts quoted per $1{,}000$ refer to it. The discovery
run is the reward of Eq.~\ref{eq:reward} without the single-element guard, as in
Fig.~\ref{fig:hack}, and the pre-creativity runs are the development runs that predate
the creativity term.

\begin{table}[h]
\caption{The reward-hacking taxonomy. Five specification-gaming modes met under
displacement, energy and $\ehull$ rewards, each closed by one targeted change.}
\label{tab:taxonomy}
\begin{center}
\footnotesize
\begin{tabular}{p{0.28cm}p{2.05cm}p{2.9cm}p{3.6cm}p{2.9cm}}
\toprule
\# & reward optimized & what the policy did & why it paid & closure \\
\midrule
1 & relaxation displacement (RMSD) & near-stationary but energetically
implausible structures & displacement carries no energy information &
energy-based reward \\
2 & formation energy & collapse onto deeply bound, far-above-hull fluorides &
strong bonding is a poor proxy for phase stability when more stable competitors
exist & switch objective to $\ehull$ \\
3 & unfloored $\ehull$ & below-hull depth chasing in sparsely referenced
regions & with few references $\ehull$ is unbounded below, and depth pays
linearly & floor at zero. $-\mathrm{clip}(\ehull,0,1)$ removes the payout by
construction \\
4 & floored $\ehull$ $+$ residual geometry term & sparse above-hull O/F
binaries gaming the geometry term & thinly referenced hulls score leniently &
route sparse-reference scores away from credit \\
5 & full recipe, abstention routing & distinct single-element packings flooding
the \msun{} set & single-element subsystems are sparsely charted, and the packings
pass the metastability, uniqueness and novelty checks of the three community
implementations we checked at source & single-element guard. A single-element
composition receives the penalty reward, and no relocation followed \\
\bottomrule
\end{tabular}
\end{center}
\end{table}

\paragraph{Mode 1, displacement.} The earliest reward scored how little a structure
moves under relaxation. With a loose relaxation the policy produced near-stationary
structures with implausible energetics. At strict force tolerance the within-group
reward spread collapsed to 0.045~\AA{}, noise level against the 1.63~eV/atom spread
of the energy-family rewards, and learning stopped.

\paragraph{Mode 2, formation energy.} A 350-rollout formation-energy run drove the
median formation energy to $-2.31$~eV/atom while the same checkpoint had the worst
hull distance of the runs compared in that study (median $\ehull$ of
$+1.08$~eV/atom). The policy collapsed onto deeply bound fluorides such as DyMgF$_2$
and Cs$_2$LiYbF$_6$. The fluorine-containing fraction of generations rose from 14\% to
91\% and the Yb-containing fraction from 1\% to 73\% between rollouts 0 and 350, and
the per-checkpoint correlation between formation energy and hull distance stayed
weak (Pearson $+0.09$ to $+0.36$).

\paragraph{Mode 3, unfloored $\ehull$.} With the objective moved to $\ehull$ but not
floored, the policy chased below-hull depth where references are thin, drifting toward
Li--O compositions whose hull distances sit far below zero only because the reference
hull contains little to compete against. Flooring the term at zero makes on-hull and
below-hull structures tie at the best value, so fictitious depth earns nothing.

\paragraph{Mode 4, sparse above-hull binaries.} Floored and combined with a residual
geometry term, the recipe drifted into sparse above-hull O/F binaries, and the
fluorine-containing fraction of generations rose from 11\% to 59\% over 750 rollouts.
A KL term anchoring the composition channel to the pretrained model at
$\beta = 5\times10^{-3}$ did not arrest the drift (held-out fluorine share 12\% to
60\%). Routing structures with fewer than 12 near-hull references away from credit
closed it. The held-out fluorine share fell from 12\% to 7\%, the sparse O/F binaries
were replaced by dense Cu- and Ag-rich oxides, and the median reference count of the
checkpoint held near 32 where the ungated run's fell to 3. This is why the guards live
in the reward rather than in a trust region.

\begin{figure}[htbp]
\centering
\includegraphics[width=0.85\linewidth]{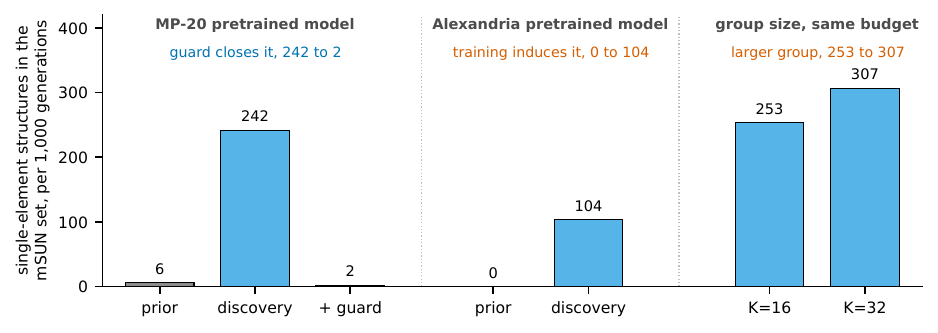}
\caption{\textbf{Where the single-element structures come from.} Single-element
structures in the \msun{} set per $1{,}000$ generations under the internal evaluation.
On the MP-20 prior the guard reduces the count from 242 to 2. The same training
induces these structures from zero on a second model pretrained on Alexandria, and on
the MP-20 model a larger group size at the same rollout budget produces more of them.
The Alexandria and group-size comparisons were run before the creativity term was
added to the reward.}
\label{fig:elemental}
\end{figure}

\paragraph{Mode 5, single-element packings.} The abstention form of that routing
created the unpunished region described in Section~\ref{sec:hack}, and
Table~\ref{tab:elemruns} and Fig.~\ref{fig:elemental} collect the counts. The pretrained prior places 6
single-element structures per $1{,}000$ generations in the \msun{} set, the
pre-creativity discovery recipe 253 and the discovery run of the main text 242. The
guard brings this to 12 in the pre-creativity run, 2 to 15 across three seeds, and to
2 for OMatGRPO. The mode is induced from zero on the roughly $20\times$ larger
Alexandria prior, which places no single-element structure in its \msun{} set on its
own and 104 per $1{,}000$ under the pre-creativity recipe, and it grows with group size
at matched rollout budget ($K{=}16$: 253, $K{=}32$: 307). Under the guard the
single-element fraction of generations holds at a mean of 0.008, its largest
excursion (0.27) is punished and self-corrects, and a pre-registered relocation watch
(an alert if the largest sparse-compound element family crossed 15) did not fire. The
residual sparse-compound \msun{} set (103 structures) is led by alkali-halide and
chalcogenide binaries, its largest family (Br--Cs) reaches 11, and its novelty against
the Alexandria reference is 0.884 (pre-creativity discovery run: 0.760).

\begin{table}[h]
\caption{Single-element structures in the \msun{} set. Upper block, per $1{,}000$
generations under the internal evaluation. Lower block, raw counts in released sample
sets of generators that were sampled rather than trained against the metric.}
\label{tab:elemruns}
\begin{center}
\footnotesize
\begin{tabular}{p{5.2cm}rp{4.0cm}}
\toprule
run & single-element in \msun{} & note \\
\midrule
pretrained prior (MP-20) & 6 & \\
discovery recipe, pre-creativity, $K{=}16$ & 253 & abstention routing, no single-element guard \\
discovery recipe, pre-creativity, $K{=}32$ & 307 & matched rollout budget \\
$+$ single-element guard, pre-creativity & 12 & 2 to 15 across three seeds \\
discovery run (Eq.~\ref{eq:reward} without the guard) & 242 & \\
OMatGRPO (Eq.~\ref{eq:reward}) & 2 & \\
penalty routing, pre-creativity & 1 to 2 & whole sparse tier penalized \\
Alexandria prior & 0 & 1 single-element structure per $1{,}000$ generations, outside the \msun{} set \\
Alexandria prior $+$ discovery recipe, pre-creativity & 104 & induced from zero \\
\midrule
MatterGen, released base model & 0 of $1{,}024$ & passive sampling \\
Chemeleon2, released MP-20 set & 87 of $10{,}000$ & latent-space RL, 28 copies of one K$_{20}$ packing \\
Chemeleon2, released Alexandria set & 24 of $10{,}000$ & latent-space RL \\
\bottomrule
\end{tabular}
\end{center}
\end{table}

\paragraph{Cross-protocol confirmation.} The first twenty offending packings, pushed
through MatterGen's released evaluation pipeline unmodified, come back 16 of 20
stable, unique and novel. Under LeMat-GenBench, 53.5\% of the discovery run's \msun{}
set is single-element (Fig.~\ref{fig:hack}a) and 49.7\% for the pre-creativity recipe
(Table~\ref{tab:allruns}), and Chemeleon2's \msun{} definition accepts the packings as
well.

\paragraph{One collapse that is not gaming.} Removing the coverage bonus produces
collapse rather than an exploit. Without it the element palette narrows to a Shannon
entropy of roughly 1.7 over emitted elements, against at least 3 held across all 750
rollouts with the bonus active, which is what motivates the diversity terms of
Eq.~\ref{eq:reward}.

%% file: appendix_c.tex

\section{Reward-swap details}\label{app:swap}

\paragraph{Chemeleon2's reward in our pipeline (B, B$'$).} The four published
Chemeleon2 components are imported unchanged and summed. These are the creativity term
(weight 1), the energy term (weight 1, single-point MACE-MP energy), the
structure-diversity term (weight 0.1) and the composition-diversity term (weight 1),
each min--max normalized over the scored batch with the MP-20 reference. Their per-batch
standardization is dropped because our GRPO advantage step supplies it, and our own
creativity and diversity terms are disabled, so the transplanted reward is their full
term set and nothing else. B scores the decoded geometry, as their reward does
natively. B$'$ relaxes each batch before scoring (FIRE, cell included, force tolerance
0.05~eV/\AA, 100 steps).

\paragraph{Our reward in Chemeleon2's pipeline (A, A$'$).} Our full reward of
Eq.~\ref{eq:reward}, with the routing of Appendix~\ref{app:config}, is installed as a
single custom component in Chemeleon2's plug-in reward slot. A flag-gated patch ports
our abstention semantics into their per-group standardization, and with the flag off
their code is byte-identical to the release. A relaxes before scoring. A$'$ removes
the relaxation, matching their pipeline's native convention.

\paragraph{Held fixed.} Each transplant keeps the host pipeline's recipe unchanged.
Our pipeline uses the configuration of Table~\ref{tab:hparams}. Their pipeline uses
their released MP-20 RL configuration, with clip 0.001, KL weight 1.0, entropy weight
$10^{-5}$, AdamW at $10^{-5}$, up to $5{,}000$ steps and the checkpoint with the best
validation reward. A$'$ ran the full horizon without early stopping. Every run
generates $2{,}500$ structures, relaxes them, and scores them with the pinned
LeMat-GenBench install of Table~\ref{tab:master}. Every run is single-seed, and A and
A$'$ generation used Chemeleon2's sampler, which is unseeded and not
bit-reproducible.

\paragraph{Results.} Table~\ref{tab:appC-full} reports every run and
Table~\ref{tab:appC-companion} the compositional and energetic companions. The
relaxation convention flips sign across pipelines. Adding relaxation before scoring
raises Chemeleon2's reward in our pipeline from 0.160 to 0.186 (B to B$'$) and lowers
our reward in their pipeline from 0.437 to 0.235 (A$'$ to A), which is why
Table~\ref{tab:swap} holds each host pipeline's convention fixed so that a swap changes
only the reward. A$'$ is the one run whose \msun{} mass shifts to quaternary and
quinary chemistry, with 4.8\% binary, 21.5\% ternary, 43.0\% quaternary, 22.6\%
quinary and 8.1\% six or more elements among its $1{,}093$ \msun{} structures, and no
single-element structures.

\begin{table}[t]
\centering
\caption{Full per-run LeMat-GenBench results. Rate columns are on the valid set
($n_{\text{val}}$), and \emph{valid} itself is on the scored set. \msun{} is given
as count, rate on valid, and rate on the $2{,}500$ nominal draw.}
\label{tab:appC-full}
\small
\setlength{\tabcolsep}{4.5pt}
\renewcommand{\arraystretch}{1.12}
\begin{tabular}{@{}lrccccrccc@{}}
\toprule
 & & \multicolumn{4}{c}{rate on valid} & & \multicolumn{3}{c}{\msun} \\
\cmidrule(lr){3-6}\cmidrule(lr){8-10}
run & $n_{\text{val}}$ & valid & uniq & novel & meta & SUN & count & /val & /2500 \\
\midrule
prior (no RL)               & 2352 & 0.948 & 0.995 & 0.821 & 0.272 & 8  & 336  & 0.143 & 0.134 \\
\addlinespace[1pt]
ours $\times$ ours          & 1745 & 0.703 & 0.987 & 0.883 & 0.758 & 7  & 1138 & 0.652 & 0.455 \\
theirs $\times$ theirs      & 2372 & 0.949 & 0.906 & 0.681 & 0.675 & 36 & 915  & 0.386 & 0.366 \\
\addlinespace[1pt]
B (theirs $\times$ ours)    & 2305 & 0.928 & 0.990 & 0.812 & 0.321 & 3  & 401  & 0.174 & 0.160 \\
B$'$ (B $+$ relax)          & 2322 & 0.933 & 0.992 & 0.817 & 0.341 & 6  & 465  & 0.200 & 0.186 \\
A (ours $\times$ theirs)    & 2268 & 0.907 & 0.997 & 0.843 & 0.338 & 6  & 587  & 0.259 & 0.235 \\
A$'$ (A $-$ relax)          & 2468 & 0.987 & 0.943 & 0.717 & 0.693 & 52 & 1093 & 0.443 & 0.437 \\
\bottomrule
\end{tabular}
\end{table}

\begin{table}[t]
\centering
\caption{Compositional and energetic companions per run. Single-element and
sparse-hull shares are over each run's \msun{} set, and dominant arity is the most
common element count and its share. Means over the valid set, energies in eV/atom,
relaxation RMSE in \AA.}
\label{tab:appC-companion}
\small
\setlength{\tabcolsep}{4pt}
\renewcommand{\arraystretch}{1.12}
\begin{tabular}{@{}lccccrr@{}}
\toprule
run & single-element in \msun & sparse in \msun & dominant arity & $\overline{E_\mathrm{hull}}$ & $\overline{E_\mathrm{form}}$ & relax RMSE \\
\midrule
prior (no RL)          & 2.1\% & 21.7\% & 3 (50\%) & 0.192 & $-0.733$ & 0.441 \\
ours $\times$ ours     & 0.4\% & 23.5\% & 3 (46\%) & 0.070 & $-0.055$ & 0.278 \\
theirs $\times$ theirs & 1.0\% & 7.9\%  & 3 (48\%) & 0.078 & $-0.804$ & 0.168 \\
B                      & 7.5\% & 26.7\% & 3 (44\%) & 0.178 & $-0.735$ & 0.456 \\
B$'$                   & 4.1\% & 22.2\% & 3 (45\%) & 0.161 & $-0.689$ & 0.448 \\
A                      & 0.0\% & 15.3\% & 3 (49\%) & 0.143 & $-0.779$ & 0.379 \\
A$'$                   & 0.0\% & 2.1\%  & 4 (43\%) & 0.070 & $-0.837$ & 0.139 \\
\bottomrule
\end{tabular}
\end{table}

\paragraph{Interpretation of the two off-diagonal cells.} This paragraph is
interpretation, and the tables above are the measurements. Chemeleon2's reward
normalizes every term onto $[0,1]$ per batch although the raw spread of its energy
term is roughly $60\times$ that of its composition-diversity term, and a replay
diagnostic shows that only about 30\% of its within-group learnable signal sits in the
energy term. Most of its gradient therefore pushes novelty and coverage rather than
hull proximity, and a metric gated on stability penalizes that allocation, which is
consistent with its weak showing in our pipeline. Our reward keeps the stability and
creativity terms on their absolute scales and rescales only the diversity bonus. In
the frozen-VAE pipeline the decoder already guarantees geometric sanity, since A$'$'s
outputs move only 0.139~\AA{} under evaluation relaxation against 0.278~\AA{} for
OMatGRPO, so a reward's entire steering budget goes to composition, and ours spends
it on the stability channel that \msun{} gates on. The traits of A$'$ agree with this
account. Its sparse-hull share falls to 2.1\%, its mean formation energy is the
deepest of any run ($-0.837$~eV/atom), and it has the highest strict-SUN count in the
study (52). The same reward produces none of these traits in our own pipeline, so the
effect is an interaction between reward and decoder rather than a property of the
reward alone. Replaying the A and A$'$ checkpoints through the same diagnostic with
and without the relaxation step would test the account, and a seed replicate of A$'$
would bound the sensitivity of its value.

\paragraph{Did our pipeline hack the transplanted reward?} We audited the B and B$'$
runs for reward rising while the quality it proxies stagnates.
Table~\ref{tab:appC-hackaudit} compares the reward's composition under the untrained
and the trained policy. Total reward climbed about 17\%, mostly through the
composition-diversity term rather than the energy term that \msun{} gates on, and
metastability moved from 0.272 to 0.321, in proportion to the small energy-term
movement. Creativity sat at its ceiling from the first step (uniqueness 1.000,
novelty 0.984 at batch scale), so it contributed no within-group variance and steered
nothing, and advantage statistics stayed well conditioned throughout. The
transplanted reward was therefore optimized without an exploit, but weakly, and with
one mispricing. B's \msun{} set is 7.5\% single-element structures, the highest share
of any run, because Chemeleon2's reward carries no term for the number of elements.
Its frozen decoder rarely decodes single-element cells (1.0\% of its own \msun{}
set), while our generator writes the composition directly and can reach them.
Relaxing before scoring roughly halves the leak (B$'$, 4.1\%). This is the
counterpart, in the other direction, of the failure mode of Section~\ref{sec:hack},
where a reward does not specify what its new host is free to vary.

\begin{table}[h]
\centering
\caption{Reward composition of the B run under the untrained and trained policy
(post-normalization means).}
\label{tab:appC-hackaudit}
\small
\begin{tabular}{@{}lccc@{}}
\toprule
term & untrained policy & trained policy & $\Delta$ \\
\midrule
creativity            & 1.000 & 0.994 & $\approx 0$ \\
composition diversity & 0.473 & 0.648 & $+0.175$ \\
energy                & 0.043 & 0.125 & $+0.082$ \\
structure diversity   & 0.035 & 0.044 & $+0.009$ \\
\midrule
total                 & 1.551 & 1.811 & $+0.260$ \\
\bottomrule
\end{tabular}
\end{table}

%% file: appendix_dft.tex

\section{DFT protocol and spot-check tables}\label{app:dft}

\paragraph{Protocol.} Structures were drawn from the relaxed generation sets of the
internal evaluation (Appendix~\ref{app:internal}) in three batches of 28, 32 and 30,
with the sampling frame written down before any calculation ran. The batches cover the
pre-creativity discovery run, anchors from the prior, the two sparse oxygen--fluorine
binaries of the mode 4 run of Appendix~\ref{app:hacks}, the guarded pre-creativity
run, probes of fluorine- and oxygen-bearing compositions, sparse-hull compounds, five
single-element packings and, in the third batch, OMatGRPO with its trusted tier
stratified by the number of elements and its sparse tier sampled by family and at
random. Within each stratum the structures were drawn at random with a fixed seed,
except for the sparse-family anchors of the third batch, which were chosen by hand to
probe the largest sparse families. Each calculation starts from the UMA-relaxed
geometry and runs in VASP 6.4.1 \citep{kresse1996} with the projector augmented-wave method, a relaxation
of positions and cell with pymatgen's \texttt{MPRelaxSet} (PBE \citep{pbe1996}, with the Materials
Project Hubbard $U$ values where they apply) followed by a static calculation with
\texttt{MPStaticSet}. The DFT energy above the hull is measured against the DFT split
of the LeMat-Bulk hull released with LeMat-GenBench, with the reference count defined
as in the reward. The Materials Project 2020 corrections \citep{wang2021corrections} are applied to neither side.
Applying them to our entries alone shifts fluorine- and oxygen-bearing structures by
0.3 to 0.5~eV/atom and leaves intermetallics untouched, and the correction status of
the LeMat-Bulk energies cannot be verified from the released files, so leaving both
sides uncorrected is the only choice that cannot double-correct. The UMA comparator is
the reward's $\ehull$ on the geometry the DFT calculation starts from, and a structure
counts as trusted or sparse by the UMA reference count with the threshold of 12 of
Section~\ref{sec:reward}. Two calculations were excluded, both on Gd-bearing
structures. B$_2$Gd$_2$Ru$_2$, a prior anchor, did not converge electronically in the
static calculation, and Gd$_4$Lu$_1$Sm$_1$, a sparse OMatGRPO structure, diverged from
the second ionic step with a final energy of $+1.7\times10^{4}$~eV although the
relaxation log reported convergence. The Materials Project settings map Gd to the
f-in-valence pseudopotential while every other lanthanide in the set uses the
f-in-core variant. The two rows enter no statistic and were not repeated with another
pseudopotential, which would break the protocol identity across batches.

\paragraph{Summary.} Table~\ref{tab:dft-tiers} gives the agreement between UMA and
DFT hull distances by tier and by run. On the trusted tier the mean absolute deviation
is 0.016~eV/atom with a bias of $+0.010$~eV/atom, DFT being the higher, and 63 of 67
metastability claims agree at the 0.1~eV/atom threshold. On the sparse tier the
energies track as well (0.020~eV/atom, 19 of 21), but the DFT hull is built from the
same reference data and is missing the same competing phases, so this agreement
validates the energies and not the stability of those compositions. Four trusted-tier
claims flip. Mg$_2$Yb$_4$ (guarded pre-creativity run) moves from $-0.012$~eV/atom
under UMA to $+0.155$~eV/atom under DFT, the largest deviation among the four, and Yb is
an element whose pseudopotential differs between the databases behind the reference
hull. Cu$_2$S$_4$Sn$_2$ (prior, $0.099$ to $0.110$~eV/atom) and Lu$_1$Pr$_2$Sm$_5$
(OMatGRPO, $0.043$ to $0.102$~eV/atom) straddle the threshold, and the latter has only
four reference phases on the DFT hull although its subsystem counts as trusted on the
UMA hull. F$_3$O$_2$Rb$_1$ (OMatGRPO, fluorine and oxygen probe) moves from $0.079$ to
$0.190$~eV/atom. The two sparse-tier flips are Ge$_{14}$Yb$_6$ ($0.095$ to
$0.149$~eV/atom), again a Yb case, and the single-element packing Rb$_6$ ($0.061$ to
$0.186$~eV/atom), whose subsystem has one reference phase. One agreement is nominal. Co$_{10}$Ho
(OMatGRPO) counts as agreeing only because its UMA value of $0.101$~eV/atom sits just above
the threshold, while DFT gives $0.283$~eV/atom, the largest deviation in the set.

\paragraph{Second potential.} The 0.015~eV/atom of Section~\ref{sec:soft} is the median
absolute difference per atom between UMA and MatterSim \citep{mattersim} energies on the
UMA-relaxed structures of the mode 4 run of Appendix~\ref{app:hacks} under the internal
evaluation (990 of its $1{,}000$ generations), after a composition-linear offset fitted by
least squares is removed, so that differences in the elemental reference energies of the two
potentials do not enter. It was computed on that development run and not repeated for
OMatGRPO.

\begin{table}[h]
\caption{UMA against DFT hull distances by tier and by run. Agreement counts
metastability claims on the same side of 0.1~eV/atom under both methods.}
\label{tab:dft-tiers}
\begin{center}
\small
\begin{tabular}{llrcc}
\toprule
tier & run & $n$ & MAE (eV/atom) & agreement \\
\midrule
trusted & prior & 4 & 0.005 & 3 of 4 \\
trusted & discovery, pre-creativity & 28 & 0.009 & 28 of 28 \\
trusted & guarded, pre-creativity & 13 & 0.019 & 12 of 13 \\
trusted & OMatGRPO & 22 & 0.026 & 20 of 22 \\
\textbf{trusted} & \textbf{all} & \textbf{67} & \textbf{0.016} & \textbf{63 of 67} \\
\midrule
sparse & discovery, pre-creativity & 10 & 0.033 & 8 of 10 \\
sparse & mode 4 run of Appendix~\ref{app:hacks} & 2 & 0.014 & 2 of 2 \\
sparse & guarded, pre-creativity & 2 & 0.009 & 2 of 2 \\
sparse & OMatGRPO & 7 & 0.007 & 7 of 7 \\
\textbf{sparse} & \textbf{all} & \textbf{21} & \textbf{0.020} & \textbf{19 of 21} \\
\bottomrule
\end{tabular}
\end{center}
\end{table}

\paragraph{Per-structure values.} Table~\ref{tab:dft-all} lists every calculation
that entered a statistic. The run codes are P for the prior, D$^{-}$ and G$^{-}$ for
the pre-creativity discovery and guarded runs, M4 for the mode 4 run and O for
OMatGRPO.

\begin{table}[htbp]
\caption{Per-structure UMA and DFT hull distances (eV/atom) for the 88 spot-checked
structures. Tier T is trusted and S is sparse by the UMA hull. Refs is the number of
reference phases within 1~meV/atom of the DFT hull in the structure's chemical
subsystem. Agree (y or n) marks metastability claims on the same side of 0.1~eV/atom.}
\label{tab:dft-all}
\begin{center}
\scriptsize
\setlength{\tabcolsep}{2.5pt}
\input{dft_table_rows}
\end{center}
\end{table}

%% file: dft_table_rows.tex
\begin{tabular}{@{}llcrrrc@{}}
\toprule
run & formula & tier & refs & UMA & DFT & agree \\
\midrule
D$^{-}$ & Ac$_{3}$Hg$_{5}$ & T & 9 & 0.054 & 0.050 & y \\
D$^{-}$ & Ac$_{9}$In$_{3}$ & T & 6 & 0.080 & 0.084 & y \\
D$^{-}$ & AgLi$_{8}$ & T & 11 & 0.007 & 0.013 & y \\
D$^{-}$ & AgLi$_{9}$ & T & 11 & 0.040 & 0.043 & y \\
D$^{-}$ & AgMg$_{9}$ & T & 7 & 0.068 & 0.085 & y \\
D$^{-}$ & AgSr$_{9}$ & T & 12 & 0.088 & 0.024 & y \\
D$^{-}$ & AgZn$_{19}$ & T & 7 & 0.070 & 0.079 & y \\
D$^{-}$ & Ag$_{3}$Li$_{5}$ & T & 11 & 0.007 & 0.011 & y \\
D$^{-}$ & Ag$_{4}$Li & T & 11 & 0.001 & 0.005 & y \\
D$^{-}$ & Ba$_{7}$Na$_{3}$ & T & 4 & 0.053 & 0.052 & y \\
D$^{-}$ & Cd$_{4}$Li$_{2}$ & T & 12 & 0.015 & 0.028 & y \\
D$^{-}$ & ClCu & T & 5 & 0.100 & 0.098 & y \\
D$^{-}$ & Hg$_{6}$Na & T & 13 & 0.032 & 0.034 & y \\
D$^{-}$ & La$_{9}$Pb$_{3}$ & T & 13 & 0.078 & 0.096 & y \\
D$^{-}$ & Li$_{11}$Mg$_{3}$ & T & 5 & 0.005 & 0.017 & y \\
D$^{-}$ & Li$_{12}$Mg$_{4}$ & T & 5 & 0.051 & 0.060 & y \\
D$^{-}$ & Mg$_{8}$Sc$_{2}$ & T & 7 & 0.031 & 0.032 & y \\
D$^{-}$ & Na$_{3}$Pb$_{7}$ & T & 8 & 0.074 & 0.084 & y \\
D$^{-}$ & Ni$_{4}$Zn & T & 8 & 0.066 & 0.067 & y \\
D$^{-}$ & Pt$_{2}$Sn$_{16}$ & T & 12 & 0.073 & 0.095 & y \\
P & BaF$_{6}$Si & T & 15 & 0.026 & 0.025 & y \\
P & CuSe$_{2}$Tm & T & 24 & 0.064 & 0.058 & y \\
D$^{-}$ & Ag$_{3}$Li$_{4}$ & T & 11 & 0.016 & 0.020 & y \\
D$^{-}$ & CdHg$_{16}$Ho & T & 31 & 0.056 & 0.055 & y \\
D$^{-}$ & CdLi$_{9}$ & T & 12 & 0.045 & 0.046 & y \\
D$^{-}$ & Hg$_{7}$Li & T & 12 & 0.025 & 0.021 & y \\
D$^{-}$ & Li$_{4}$Mg & T & 5 & 0.002 & 0.010 & y \\
D$^{-}$ & NaO$_{4}$ & T & 8 & 0.030 & 0.009 & y \\
D$^{-}$ & Na$_{2}$O$_{7}$ & T & 8 & 0.044 & 0.030 & y \\
D$^{-}$ & O$_{11}$Rb & T & 8 & 0.056 & 0.047 & y \\
G$^{-}$ & AgGa$_{15}$ & T & 4 & 0.056 & 0.054 & y \\
G$^{-}$ & Ag$_{4}$Ga$_{16}$ & T & 4 & 0.053 & 0.061 & y \\
G$^{-}$ & Al$_{3}$Sm$_{8}$Y$_{2}$ & T & 15 & 0.089 & 0.086 & y \\
G$^{-}$ & AuMnPd & T & 27 & 0.050 & 0.036 & y \\
G$^{-}$ & Cd$_{5}$Hg$_{2}$Li$_{2}$ & T & 30 & 0.045 & 0.048 & y \\
G$^{-}$ & Cu$_{12}$MnPd & T & 20 & 0.038 & 0.032 & y \\
G$^{-}$ & K$_{2}$Rb$_{2}$Sb$_{2}$ & T & 18 & 0.021 & 0.020 & y \\
G$^{-}$ & Mg$_{2}$Yb$_{4}$ & T & 5 & -0.012 & 0.155 & n \\
G$^{-}$ & Na$_{3}$Sb$_{3}$ & T & 6 & 0.090 & 0.089 & y \\
G$^{-}$ & O$_{16}$RbSr & T & 12 & 0.080 & 0.072 & y \\
G$^{-}$ & O$_{7}$Rb & T & 8 & 0.075 & 0.078 & y \\
G$^{-}$ & O$_{9}$Rb & T & 8 & 0.067 & 0.079 & y \\
G$^{-}$ & Pd$_{2}$Zr$_{2}$ & T & 15 & 0.047 & 0.033 & y \\
P & As$_{5}$BaCu$_{7}$ & T & 21 & 0.070 & 0.067 & y \\
\bottomrule
\end{tabular}
\hfill
\begin{tabular}{@{}llcrrrc@{}}
\toprule
run & formula & tier & refs & UMA & DFT & agree \\
\midrule
P & Cu$_{2}$S$_{4}$Sn$_{2}$ & T & 17 & 0.099 & 0.110 & n \\
O & AcLaNdTm & T & 6 & 0.050 & 0.056 & y \\
O & AlErPr$_{9}$Sm & T & 31 & 0.052 & 0.065 & y \\
O & CdMgTl$_{16}$ & T & 17 & 0.021 & 0.045 & y \\
O & Co$_{10}$Ho & T & 15 & 0.101 & 0.283 & y \\
O & CsCu$_{4}$F$_{15}$ & T & 23 & 0.079 & 0.068 & y \\
O & CsO$_{2}$Rb & T & 17 & 0.012 & 0.004 & y \\
O & Cu$_{3}$Zn$_{17}$ & T & 8 & 0.036 & 0.050 & y \\
O & ErMg$_{9}$Y & T & 15 & 0.035 & 0.042 & y \\
O & ErNd$_{7}$SmTm & T & 15 & 0.015 & 0.021 & y \\
O & Er$_{2}$Sc$_{8}$SmTm & T & 10 & 0.027 & 0.087 & y \\
O & F$_{3}$O$_{2}$Rb & T & 17 & 0.079 & 0.190 & n \\
O & GaLa$_{3}$Pr & T & 29 & 0.077 & 0.088 & y \\
O & GaTm$_{4}$Y & T & 22 & 0.074 & 0.073 & y \\
O & HgK$_{19}$ & T & 14 & 0.026 & 0.037 & y \\
O & HoMgTmY$_{9}$ & T & 25 & 0.049 & 0.054 & y \\
O & HoNdPrTb$_{5}$ & T & 17 & 0.086 & 0.092 & y \\
O & Li$_{5}$Zn & T & 8 & 0.011 & 0.013 & y \\
O & LuPr$_{2}$Sm$_{5}$ & T & 4 & 0.043 & 0.102 & n \\
O & LuSmY$_{3}$ & T & 6 & 0.030 & 0.033 & y \\
O & Na$_{3}$Sn$_{8}$ & T & 9 & 0.069 & 0.073 & y \\
O & Nd$_{11}$Ni$_{5}$ & T & 11 & 0.088 & 0.097 & y \\
O & Nd$_{8}$TmY & T & 8 & 0.038 & 0.048 & y \\
D$^{-}$ & Cs$_{9}$Na$_{3}$ & S & 4 & 0.053 & 0.082 & y \\
D$^{-}$ & Cu$_{12}$ & S & 2 & 0.007 & 0.003 & y \\
D$^{-}$ & O$_{4}$ & S & 1 & 0.028 & 0.011 & y \\
M4 & F$_{11}$O$_{3}$ & S & 6 & 0.073 & 0.087 & y \\
M4 & F$_{17}$O$_{3}$ & S & 6 & 0.073 & 0.086 & y \\
D$^{-}$ & As$_{4}$ & S & 1 & 0.030 & 0.021 & y \\
D$^{-}$ & F$_{8}$ & S & 1 & -0.003 & 0.019 & y \\
D$^{-}$ & Ge$_{14}$Yb$_{6}$ & S & 5 & 0.095 & 0.149 & n \\
D$^{-}$ & Ho$_{2}$Nd$_{12}$ & S & 4 & 0.013 & 0.032 & y \\
D$^{-}$ & Li$_{3}$ & S & 1 & 0.059 & 0.013 & y \\
D$^{-}$ & Rb$_{6}$ & S & 1 & 0.061 & 0.186 & n \\
D$^{-}$ & Sr$_{16}$ & S & 2 & 0.013 & 0.007 & y \\
G$^{-}$ & Ca$_{7}$Cd$_{2}$ & S & 9 & 0.031 & 0.034 & y \\
G$^{-}$ & Ce$_{13}$Rh$_{2}$ & S & 6 & 0.080 & 0.096 & y \\
O & Au$_{2}$Cd$_{10}$ & S & 8 & 0.039 & 0.037 & y \\
O & BaLuSm$_{8}$ & S & 5 & 0.086 & 0.097 & y \\
O & Ca$_{3}$Mg$_{11}$ & S & 4 & 0.080 & 0.092 & y \\
O & CdK$_{5}$ & S & 7 & 0.036 & 0.045 & y \\
O & K$_{11}$O$_{9}$ & S & 7 & 0.084 & 0.079 & y \\
O & K$_{12}$Pb & S & 6 & 0.044 & 0.041 & y \\
O & LiNd$_{7}$Tm & S & 5 & 0.062 & 0.072 & y \\
\bottomrule
\end{tabular}

%% file: appendix_config.tex

\clearpage
\section{Training configuration and best-of-$N$ definition}\label{app:config}

\paragraph{Model, sampler and optimization.} All runs start from the OMatG checkpoint
pretrained on MP-20 for de novo generation \citep{omatg}, which is also the reference
policy of the KL terms. Rollouts use a time grid of 64 points on the unit interval, so
63 Euler steps, with species noise $\eta=0$ and the velocity annealing of the
pretrained sampler switched off. Each iteration draws $B=4$ groups of $K=16$
structures, each group from one draw of the initial noise and masks, and scores them
with Eq.~\ref{eq:reward}. Within a group the rewards are clipped to three standard
deviations around the group mean, the mean and standard deviation are taken over the
trusted members only, that is over the structures not routed to abstention, and every
member is normalized with these two numbers. Members routed to abstention receive
advantage zero, and so does every member of a group with fewer than two trusted
members or a standard deviation below $10^{-4}$. The batch is then used for three
inner epochs of Eq.~\ref{eq:grpo} with AdamW, learning rate $10^{-4}$ and
gradient-norm clipping at 0.5, so 750 rollout iterations are $2{,}250$ optimizer
steps. Table~\ref{tab:hparams} lists the remaining values, and one run takes about
eight hours on one NVIDIA B200 GPU. The channel weights $\alpha_f$ were set from the
logged magnitudes of the per-channel surrogate terms in short calibration runs. At
equal weights the species term is far smaller than the position term, since it
averages single categorical log-probabilities over commit events while the position
term averages the log-density of all atoms over integration steps, so the species
channel carries weight 1 and the two continuous channels 0.1.

\begin{table}[h]
\caption{Hyperparameters of the OMatGRPO run. The discovery run of
Section~\ref{sec:hack} uses the same values with the single-element guard removed.}
\label{tab:hparams}
\begin{center}
\small
\begin{tabular}{p{5.4cm}p{7.4cm}}
\toprule
setting & value \\
\midrule
groups per iteration $B$, group size $K$ & 4, 16 \\
time grid, species noise $\eta$ & 64 points, 0 \\
rollout iterations, inner epochs & 750, 3 \\
optimizer, learning rate, gradient-norm clip & AdamW, $10^{-4}$, 0.5 \\
clip range $\epsilon_{\mathrm{clip}}$ of Eq.~\ref{eq:grpo} & 0.2 \\
channel weights $\alpha_{\mathrm{species}},\ \alpha_{\mathrm{pos}},\ \alpha_{\mathrm{cell}}$ & 1, 0.1, 0.1 \\
KL weights $\beta_{\mathrm{species}},\ \beta_{\mathrm{pos}},\ \beta_{\mathrm{cell}}$ & 0.05, 0.01, 0 \\
reward clip within a group & $\pm 3$ standard deviations \\
relaxation inside the reward & FIRE under UMA, cell included, at most 100 steps, force tolerance 0.05~eV/\AA \\
stability term $r_{\mathrm{stab}}$ & $-\mathrm{clip}(\ehull, 0, 1)$ \\
sparse-hull abstention & fewer than 12 reference phases within 1~meV/atom of the hull \\
floor $r_{\mathrm{pen}}$ & $-1$, for single-element structures and for $\ehull < -0.1$~eV/atom \\
creativity weight, partial credit & 1, clamped AMD distance ($k=100$) \\
occurrence discount $c_{\mathrm{occ}}$ & 1 up to 3 occurrences, 0 from 6 \\
MMD bonus & weight 0.4, polynomial kernel $(x\cdot y + 1)^3$, per-batch min--max scaling \\
\bottomrule
\end{tabular}
\end{center}
\end{table}

\paragraph{Relaxation before scoring.} The sampler's raw geometries move by about
1~\AA{} in RMSD when they are relaxed. Scored on the raw geometry, the stability term
would measure that distance rather than the energy of the phase, and the novelty term
would pay full credit to almost every structure, since at most 0.2\% of raw
geometries match a known structure against 4 to 13\% after relaxation. A term that is
nearly constant within a group carries no gradient under group-relative advantages.
Every rollout structure is therefore relaxed with at most 100 FIRE \citep{fire2006} steps, including
the cell, under UMA (UMA-s-1p2, task omat) before any term is computed.

\paragraph{Stability term and routing.} $\ehull$ is measured against the
UMA-consistent LeMat-Bulk hull released with LeMat-GenBench, and the reference count
of a structure is the number of hull entries within 1~meV/atom of the hull in its
chemical subsystem. A structure with fewer than 12 reference phases is routed to
abstention. It keeps its reward value, but its advantage is set to zero and it is left
out of the group statistics. A single-element structure receives the floor
$r_{\mathrm{pen}} = -1$ as its stability term $r_{\mathrm{stab}}$, and so does a
structure more than 0.1~eV/atom below the hull, because the hull is built from the same
potential and a value that far below it marks a failure of the potential. The creativity
term is evaluated for these structures as for any other, so their reward is at most
$r_{\mathrm{pen}} + c_{\mathrm{occ}}\, r_{\mathrm{creat}} + 0.4\, r_{\mathrm{mmd}}$, one
unit of $r_{\mathrm{stab}}$ below a structure on the hull with the same creativity credit.
The floor takes precedence over abstention, since a single-element structure is always
sparsely referenced and would otherwise abstain instead of being punished, which is
the exploit of Section~\ref{sec:hack}. A structure whose relaxation fails or whose
cell fails the sanity checks of the reward also receives the floor.

\paragraph{Creativity term.} The term follows Chemeleon2 \citep{chemeleon2rl}. A
relaxed structure is novel if no structure of the same reduced formula in the MP-20
training split matches it under the pymatgen structure matcher with default
tolerances, and unique if no earlier structure of the same reduced formula in the
rollout batch matches it. A structure that is both scores 1, one that is neither
scores 0, and one that is only novel or only unique receives the minimum
average-minimum-distance \citep{widdowson2022amd} between its relaxed geometry and
the matching structures ($k=100$), clamped to $[0, 1]$. Chemeleon2 scores novelty
against the whole MP-20 set and leaves the partial credit unclamped. A structure whose
matcher call exceeds three seconds scores 0.

\paragraph{Diversity terms.} The occurrence discount acts within each group. With
$n$ the number of structures in the group that share a composition, taken as the
multiset of species, $c_{\mathrm{occ}} = 1$ for $n \leq 3$, falls linearly to 0 at
$n = 6$, and is applied around the floor as in Eq.~\ref{eq:reward}, so a repeated
composition is pulled toward $r_{\mathrm{pen}}$ whatever the sign of its reward. The
MMD bonus acts across the whole batch of $BK = 64$ structures. Each structure is
represented by its fractional composition vector over 119 element slots, the squared
maximum mean discrepancy \citep{gretton2012mmd} between the batch and $10{,}000$
compositions drawn at random from the MP-20 training split is computed with the
polynomial kernel $k(x, y) = (x \cdot y + 1)^3$, and the credit of a structure is the
increase of that discrepancy when the structure is left out of the batch. The credits
of a batch are rescaled to $[0, 1]$ by their minimum and maximum, following
Chemeleon2, and enter Eq.~\ref{eq:reward} with weight 0.4. This rescaling is the only
normalization in the reward.

\paragraph{Frozen-composition control.} The control of Section~\ref{sec:gains} trains
only the position and lattice channels. The composition of each group is drawn once per
rollout iteration from the pretrained model and is shared by the 16 members of the group.
The reward is the stability term of Eq.~\ref{eq:reward} with the same floor, cap, sparse
abstention and relaxation step limit, with the relaxation moving the positions only. The
creativity, occurrence and MMD terms are omitted, because every member of a group shares
one composition, so the occurrence discount would penalize every structure and the MMD
bonus has nothing to steer, and the single-element guard is not needed because the
compositions are the pretrained model's. The channel weights are
$\alpha_{\mathrm{pos}} = \alpha_{\mathrm{cell}} = 1$ with $\beta_{\mathrm{pos}} = 0.01$, and the
rollout budget, sampler, optimizer and number of iterations are those of
Table~\ref{tab:hparams}.

\paragraph{Best-of-$N$ definition.} We drew $47{,}795$ structures from the pretrained
model with the training-time sampler and scored each with the stability pathway of
the training reward, a FIRE relaxation of the positions of at most 100 steps followed
by $-\mathrm{clip}(\ehull, 0, 1)$, with structures on sparse hulls, which includes
every single-element structure, and structures more than 0.1~eV/atom below the hull
routed to the floor. The structures were ranked by this score, ties were broken by
the lower raw relaxed $\ehull$ and then by sample index, and the top $2{,}500$ were
relaxed with the cell included and submitted to LeMat-GenBench like every other
method. Because the routing acts at scoring time, the selected set contains no
sparse-hull and no single-element structure by construction, which is why those
entries of Table~\ref{tab:master} are zero. The same ranking cut at $1{,}000$ gives
an \msun{} rate of 4.4\%.